\documentclass[a4paper,fleqn]{cas-dc}

\usepackage[numbers]{natbib}

\usepackage{amsmath,amsfonts}
\usepackage{algorithm}
\usepackage{bm}
\usepackage{array}
\usepackage[caption=false,font=normalsize,labelfont=sf,textfont=sf]{subfig}
\usepackage{textcomp}
\usepackage{stfloats}
\usepackage{url}
\usepackage{verbatim}
\usepackage{graphicx}
\usepackage{booktabs}
\usepackage{color}
\usepackage{setspace}
\usepackage{multirow}
\usepackage{algorithmicx}
\usepackage{algpseudocode}  
\usepackage[ruled,vlined, algo2e]{algorithm2e}
\SetKwRepeat{Do}{do}{while}%
\providecommand{\SetAlgoLined}{\SetLine}

\usepackage{graphics} 
\usepackage{epsfig} 
\usepackage{times} 
\usepackage{amssymb}  

\usepackage{pifont}       

\def\tsc#1{\csdef{#1}{\textsc{\lowercase{#1}}\xspace}}
\tsc{WGM}
\tsc{QE}
\tsc{EP}
\tsc{PMS}
\tsc{BEC}
\tsc{DE}

\begin{document}
\let\WriteBookmarks\relax
\def\floatpagepagefraction{1}
\def\textpagefraction{.001}

\shorttitle{AVS-Net}

\shortauthors{H. Yang et~al.}

\title [mode = title]{AVS-Net: Point Sampling with Adaptive Voxel Size for 3D Scene Understanding}                      



%
\author[1]{Hongcheng Yang}[style=chinese]






\affiliation[1]{organization={Huazhong University of Science and Technology},
    city={Wuhan},
    postcode={430074}, 
    country={China}}

\author[1]{Dingkang Liang}[style=chinese]



\affiliation[2]{organization={Baidu Inc.},
    city={Beijing},
    postcode={100193}, 
    country={China}}

\author[1]{Dingyuan Zhang}[style=chinese]
\author[1]{Zhe Liu}[style=chinese]
\author[2]{Zhikang Zou}[style=chinese]
\author[1]{Xingyu Jiang}[style=chinese]

\author[1]{Yingying Zhu}[style=chinese]
\cormark[1]


\cortext[cor1]{Corresponding author}
\cortext[0]{E-mail addresses: $^{a}$hcyang, dkliang, dyzhang233, zheliu1994, jiangxy998, yyzhu@hust.edu.cn;  $^{b}$zouzhikang@baidu.com}



\begin{abstract}
The recent advancements in point cloud learning have enabled intelligent vehicles and robots to comprehend 3D environments better. However, processing large-scale 3D scenes remains a challenging problem, such that efficient downsampling methods play a crucial role in point cloud learning. Existing downsampling methods either require a huge computational burden or sacrifice fine-grained geometric information. For such purpose, this paper presents an advanced sampler that achieves both high accuracy and efficiency. The proposed method utilizes voxel centroid sampling as a foundation but effectively addresses the challenges regarding voxel size determination and the preservation of critical geometric cues. 
Specifically, we propose a Voxel Adaptation Module that adaptively adjusts voxel sizes with the reference of point-based downsampling ratio. This ensures that the sampling results exhibit a favorable distribution for comprehending various 3D objects or scenes. Meanwhile, we introduce a network compatible with arbitrary voxel sizes for sampling and feature extraction while maintaining high efficiency. The proposed approach is demonstrated with 3D object detection and 3D semantic segmentation. Compared to existing state-of-the-art methods, our approach achieves better accuracy on outdoor and indoor large-scale datasets, \emph{e.g.} Waymo and ScanNet, with promising efficiency. The code will be released.
\end{abstract}




\maketitle

\section{Introduction} \label{sec:intro}

\begin{figure}[t]
    \centering
    \includegraphics[width=0.49\textwidth]{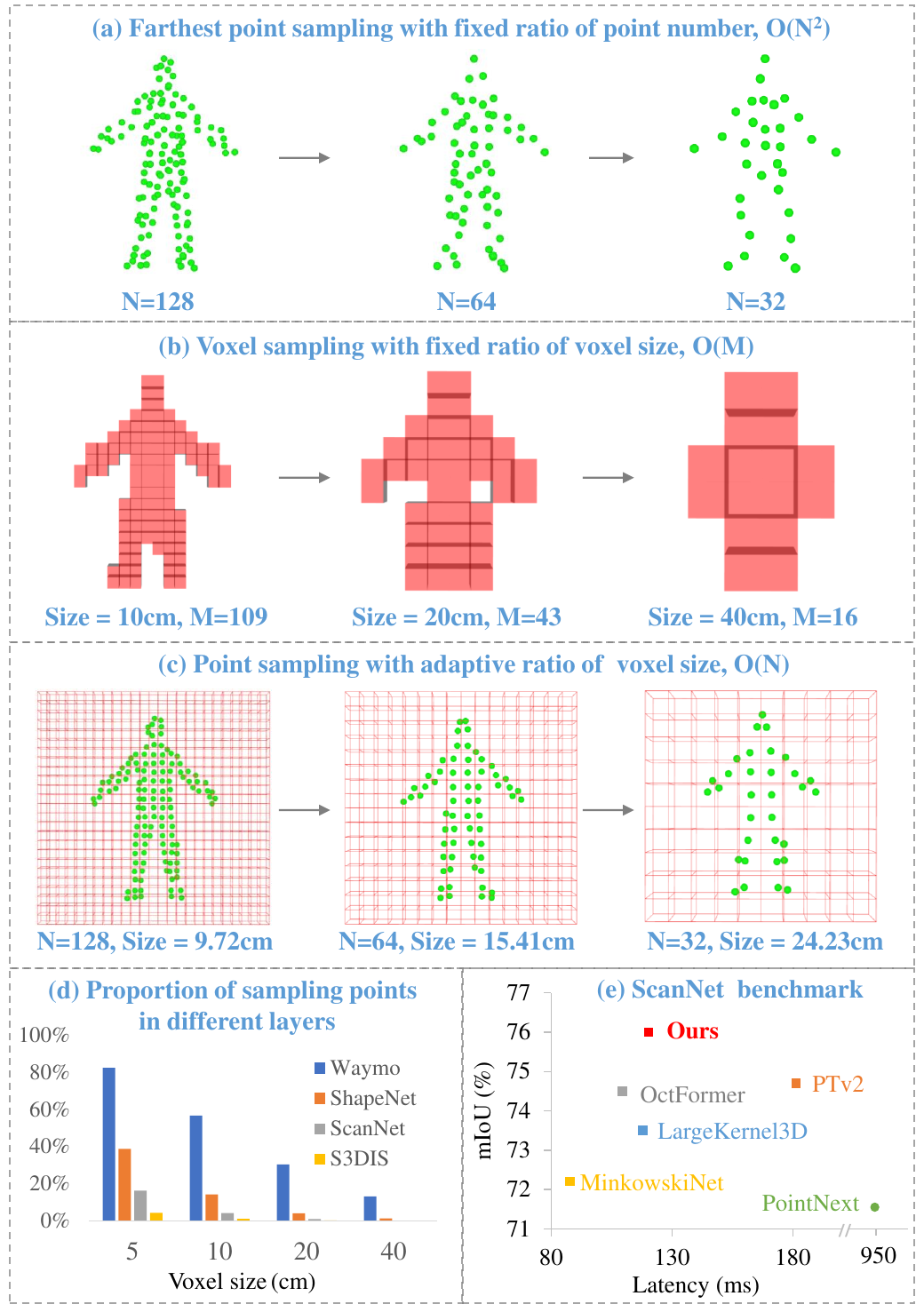}
    \caption{(a) Farthest point sampling with fixed ratio of point number, N means point number.
    (b) Voxel sampling with fixed ratio of voxel size, M means voxel number. (c) Point sampling with adaptive ratio of voxel size. (d) Proportion of sampled points in different layers. (e) Evaluation on ScanNet validation set.}\label{fig:intro0}
   \vspace{-0.5cm}
\end{figure}

Intelligent vehicles and robots need to accurately understand the 3D structure and the surrounding environment during actual motion to make appropriate decisions and plan paths~\cite{tang2023multi}.
In outdoor road scenarios, 3D object detection can assist intelligent vehicles in detecting various objects, such as vehicles, pedestrians, cyclists, etc., and provide their positions, sizes, and motion directions~\cite{sun2020scalability}.
In indoor environments, 3D semantic segmentation can segment point clouds into semantic parts, such as floors, walls, furniture, and doors. Low-speed autonomous driving robots typically operate in narrow spaces, requiring higher precision in scene understanding, making 3D semantic segmentation particularly crucial for them~\cite{dai2017scannet, park2022fast}.

The inputs often contain a large number of points in practical scenes~\cite{park2022fast,sun2020scalability}, such that the downsampling process is critical for existing learning frameworks~\cite{zhao2021point,park2022fast,chen2023largekernel3d}. Current samplers extract a representative subset from raw point clouds with point-, voxel- or learning-based designs~\cite{dovrat2019learning,wu2023attention,zhao2021point,park2022fast,zhang2023starting, chen2023largekernel3d}.
However, there still remain great challenges to develop satisfying downsampling methods that ensure both efficiency and accuracy due to the commonly existing complex geometric structures and uneven distributions in real point data.

The most intuitive and efficient strategy is random sampling~\cite{hu2020randla}. However, it easily neglects critical geometric and semantic information due to its random nature. To this end, Farthest Point Sampling (FPS) fully utilizes the distance prior for uniform sampling, thus achieving impressive accuracy and being the primary choice in most point-based networks~\cite{zhao2021point,qian2022pointnext,zhang2023starting}. 
As shown in Fig.~\ref{fig:intro0}(a), the number of sampled points is controllable~\cite{qi2017pointnet++} in FPS, making it versatile for both indoor and outdoor scenes wherein the point clouds vary a lot for different scales and ranges. However, FPS is highly time-consuming, which requires computational complexity of  $\mathcal{O}(N^2)$ for distance measuring. That means FPS may occupy most of the time in the testing process, particularly for large-scale point clouds~\cite{park2022fast}. 

As an alternative, voxel-based methods have enjoyed significant success in large-scale 3D scene understanding due to their efficient designs~\cite{park2022fast}. The original point cloud can be spatially divided into non-overlapping voxels by predefining the voxel size. Then, points within the same voxel are pooled into a single one before being fed into each network layer. Voxel sampling is quite efficient, which merely costs linear complexity, but it may leave another two problems: 
  \emph{i) To predefine the voxel size across different 3D scenes is troublesome.} As shown in Fig.~\ref{fig:intro0}(d), a fixed voxel size would produce significant differences in downsampling ratios. That means it causes under- or over-sampling for point clouds. 
  \emph{ii) Voxlization intrinsically loses fine-grained geometric structures.} As shown in Fig.~\ref{fig:intro0}(b), with the network going deeper, current methods require voxel sizes to be expanded by integer multiples (as commonly used $\times 2$), causing the voxel volume to increase by 8 times or more~\cite{choy20194d}.
Hence, the number of preserved voxels would be decreased dramatically after progressive downsamplings, as revealed in Fig.~\ref{fig:intro0}(d). This information loss often produces blurry/incorrect perceptions at object boundaries~\cite{zhang2020deep, lu2023improving}, which is the main reason for the low accuracy of existing voxel-based networks.

Additionally, researchers attempted to directly learn to downsample the raw point clouds by employing Multi-Layer Perceptrons (MLPs)~\cite{dovrat2019learning} or attention mechanisms~\cite{wu2023attention}. However, this learning paradigm involves extensive computation and memory budgets, limiting its applicability to large-scale point clouds. By contrast, the above-mentioned handcrafted designs~\cite{qi2017pointnet++,zhao2021point,zhang2023starting,choy20194d,chen2023largekernel3d} can adequately compress point clouds and are more practically used.

In this paper, we contribute to exploring an advanced sampler, namely Voxel Adaptation Module (VAM), that achieves both high efficiency and accuracy. In detail, we encourage the sampled results of our proposed VAM to mimic the good distribution of FPS. As shown in Fig.~\ref{fig:intro0}(c), our VAM automatically adjusts the voxel size to adapt to the predefined downsampling ratio across the entire dataset. With such operation, the voxel size used in each layer can be scaled at an arbitrary ratio to adapt to different density distributions, which is more favorable for preserving fine-grained features compared to a fixed scaling factor. As FPS has been widely used in various point cloud tasks, the downsampling ratio based on FPS has undergone thorough ablation studies~\cite{qi2017pointnet++,zhao2021point,qian2022pointnext,zhang2023starting}. Therefore, we can refer to the downsampling ratio used in point-based methods for configuration in practical applications.

Considering that existing learning frameworks~\cite{choy20194d, yan2018second, he2022voxel} can not directly support the feature extraction from the arbitrary-ratio voxel size, we propose the AVS-Net network. A point dynamic grouping strategy is introduced to achieve point sampling and local feature aggregation with arbitrary voxel sizes. 
We validate the superiority of our approach on 3D object detection and semantic segmentation tasks, which achieves better accuracy and promising efficiency compared to existing state-of-the-art methods, as shown in Fig.~\ref{fig:intro0}(e). Furthermore, to validate the generalizability of our method, we conducte experiments on the object part segmentation dataset ShapenetPart, achieving state-of-the-art accuracy.

Our contributions can be summarized as follows:
\begin{itemize}
\item  We introduce the Voxel Adaptation Module that automatically adjusts the voxel sizes for different scale scenes, referencing the point-based downsampling ratio. 
\item  We propose the AVS-Net network with a dynamic point grouping strategy. It supports point sampling and feature extraction with arbitrary voxel sizes while maintaining high efficiency.
\item  We validate the effectiveness of our approach to 3D object detection, semantic segmentation, and object part segmentation tasks, which achieves superior accuracy and promising efficiency compared to existing state-of-the-art methods.
\end{itemize}
\section{Related Work}
\label{sec:related}

\subsection{Point Cloud Understanding}

According to different modeling approaches for point clouds, existing 3D point cloud understanding methods can be classified into projection-based, voxel-based, and point-based networks. Projection-based methods project point clouds onto the 2D image plane using different encoding methods and then extract features using CNNs \cite{su15mvcnn, chen2017multi,lang2019pointpillars,shi2022pillarnet}. These methods are usually efficient, but the projection operation loses 3D structural details, resulting in suboptimal accuracy. Voxel-based methods convert continuous 3D point cloud into discrete voxel representations, enabling the application of 3D convolutional networks to extract features~\cite{choy20194d, yan2018second, yin2021center, chen2023voxelnext, chen2023largekernel3d}. Voxel-based methods have good computational efficiency, but voxelization loses geometric details, which is not conducive to learning fine-grained features~\cite{tang2020searching, zhang2022pvt}. Point-based Methods commonly encode and learn directly from points without voxelization or projection to 2D images~\cite{qi2017pointnet++, zhao2021point, wu2022point, shi2019pointrcnn, zhang2022not, dbqssd}. Point-based methods adopt a hierarchical grouping structure for point sets and gradually extract features of larger local areas along the hierarchy. While these methods are powerful, expensive FPS or KNN algorithms make their efficiency a bottleneck for large-scale scenes~\cite{wu2022point}.

\subsection{Voxel Sampling Methods}

Voxel sampling divides 3D space into non-overlapping uniform grids, making the points easy to search. There are two primary methods for pooling those points within the same voxel. The most direct strategy is to use the central location as represent, \emph{i.e.,} voxel center sampling~\cite{liu2019pvcnn,zhang2020deep,shi2020pv,zhang2022pvt}. 
After voxelization, deep models use integer indices of voxels for downsampling and neighbor search, as shown in Fig.~\ref{fig:intro0}(b). For multi-layer networks, voxel size can be only scaled in integer multiples, which is limited by the implementation of their networks~\cite{yan2018second,choy20194d,he2022voxel,chen2023largekernel3d}. Thus, voxel center sampling usually causes large quantization errors. 
 
Another strategy is voxel centroid sampling, which averages the coordinates of the points within the same voxel as sampling. This approach preserves the accuracy of point positions to some extent, thereby retaining fine-grained geometric information~\cite{park2022fast}. On this basis, PTv2~\cite{wu2022point}, HAVSampler~\cite{ouyang2023hierarchical} and FastPointTrans~\cite{park2022fast} use this sampling approach to ensure accuracy. However, PTv2~\cite{wu2022point} and HAVSampler~\cite{ouyang2023hierarchical} still employs expensive KNN/BallQuery for aggregating neighboring points. While FastPointTrans~\cite{park2022fast} maintains continuous point cloud coordinates, its underlying implementation still relies on MinkowskiEngine~\cite{choy20194d}, such that its downsampling is limited by the voxel size of integer multiples. For instance, it cannot perform hierarchical sampling using the voxel size $[9.72\rightarrow 15.41\rightarrow 24.23]$ in Fig.~\ref{fig:intro0}(c). 

In this paper, we develop an adaptive voxel sampling strategy to cater to various scales of 3D scenes, better preserving the fine-grained geometric relationships of point clouds. 
Additionally, to support arbitrary voxel size sampling and feature extraction, we propose a novel learning framework, AVS-Net, which achieves high performance in both accuracy and efficiency, as revealed in Fig.~\ref{fig:intro0}(e).

\section{Method}
\label{sec:method}

\subsection{Overview}
In our method, AVS-Net first utilizes VAM to determine the voxel size for each downsampling layer, and then proceeds with model training and evaluation.
For the segmentation task, AVS-Net employs a U-Net architecture that consists of an encoder and a decoder. For the detection task, AVS-Net comprises an encoder and a detection head. In the encoder part, we design the Voxel Set Abstraction (VSA) as the basic module. The details are introduced in Sec.~\ref{Sec:VSA} and shown in Fig.~\ref{fig:vsa}. Sec.~\ref{Sec:VAM} describes the adaptive voxel sampling in detail. As for the decoder, trilinear interpolation~\cite{qi2017pointnet++} is used to map the features of the downsampled subsets back to their respective supersets. 

\subsection{Voxel Set Abstraction} \label{Sec:VSA}
The VSA module consists of two steps: intra-voxel feature extraction (Intra-VFE) and inter-voxel feature extraction (Inter-VFE). 
Intra-VFE utilizes voxel centroid sampling to aggregate the features of multiple points within the same voxel onto the sampled point. Unlike other voxel-based approaches that retain only voxel indices, our approach preserves the precise coordinate information of sampled points. This is crucial for extracting fine-grained information in point cloud analysis~\cite{park2022fast,zhang2020deep,shi2020pv}.
Inter-VFE fuses features among sampled points, which can expand the receptive field of the sampled points.

\textbf{Unified data representation}.
For the arbitrary-size voxel sampling algorithm proposed in this paper, we need to address three core issues: 1) the relationship between points and voxels is many-to-one, requiring the establishment of a mapping relationship between them. 2) Aggregating multiple points, which are disorderly distributed in memory, into their corresponding voxels. 3) Forming a batch from different numbers of point clouds for parallel computation.
We use $group\_id$ for the first issue to explicitly model the many-to-one mapping relationship between points and voxels. We use a $scatter$ operator for the second issue to complete the many-to-one aggregation operation. For the third issue, we form a 2D tensor matrix from multiple frames of point clouds and use $batch\_id$ to distinguish the belonging relationships. Given a point set $M={P, F}$, each point can be expressed as $M_i = (P_i, F_i)$, where $F_i$ is the feature corresponding to the point $P_i$.
We define each point $P_i$ with the following form:
\begin{equation}
P_i = [b, x , y , z , \textit{vc\_1d}, group\_id]. 
\label{eq:data_pre}
\end{equation}

Among Eq.~\ref{eq:data_pre}, $b$ represents the $batch\_id$, used to differentiate which frame's point cloud each point belongs to. The number of points in different frames of point clouds can be unequal.
The $xyz$ represents the point's 3D spatial coordinates. The \textit{vc\_1d} is short for \textit{voxel\_coord\_1d}, representing the one-dimensional voxel coordinates obtained by merging 3D voxel coordinates. The $group\_id$ signifies the point's grouping information. In Intra-VFE, the $intra\_group\_id$ represents grouping information for multiple points belonging to the same voxel. In Inter-VFE, the $inter\_group\_id$ represents grouping information for the voxel-sampled point and its neighboring points.

\begin{figure}[t]
    \begin{minipage}{\linewidth}
        \begin{center}
        \includegraphics[width=0.85\linewidth]{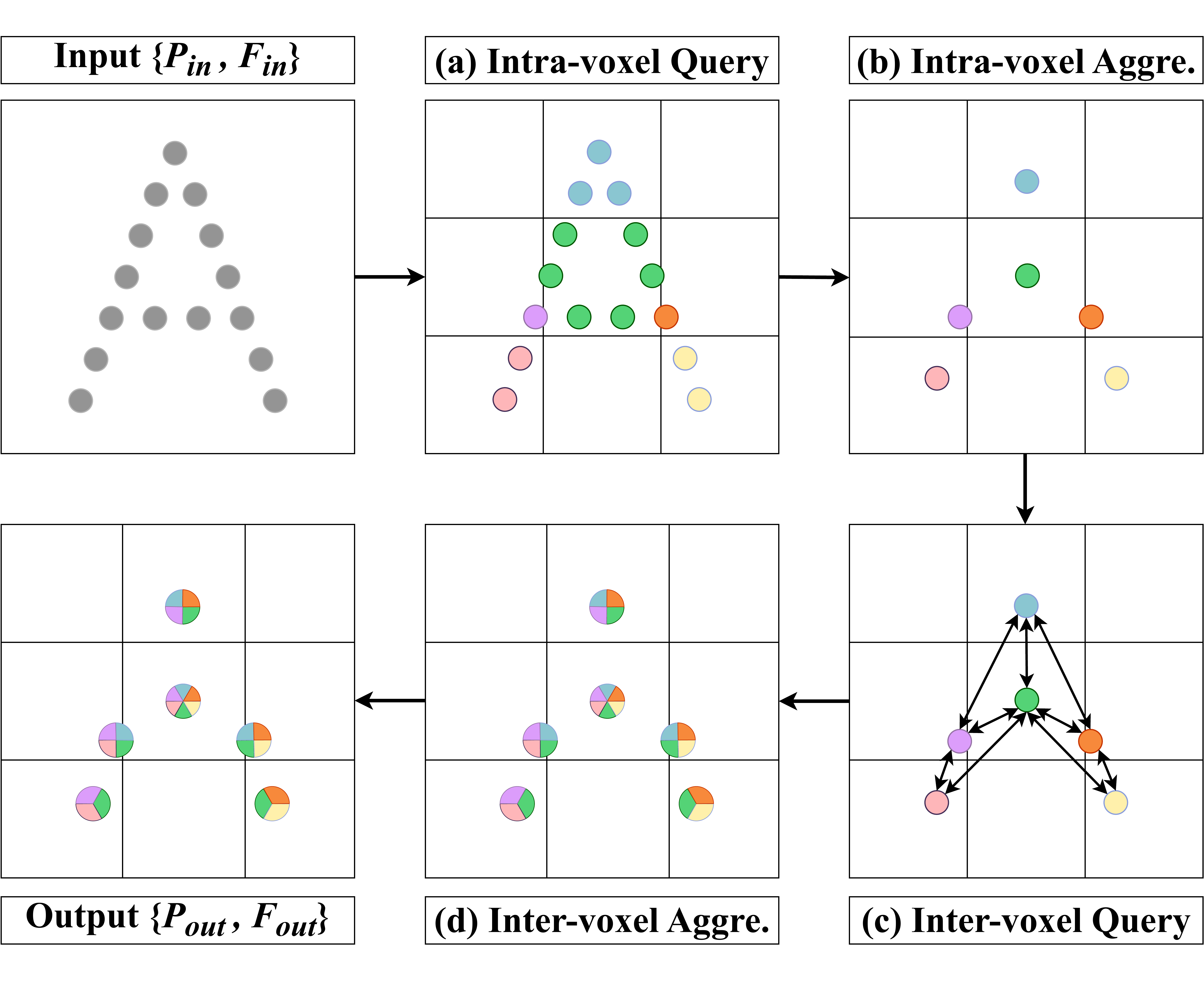}
        \caption{Schematic diagram of the Voxel Set Abstraction module}
        \label{fig:vsa}
        \end{center}
    \end{minipage}%

    \vfill 
    \begin{minipage}{\linewidth}
    \begin{center}
        \includegraphics[width=0.7\linewidth]{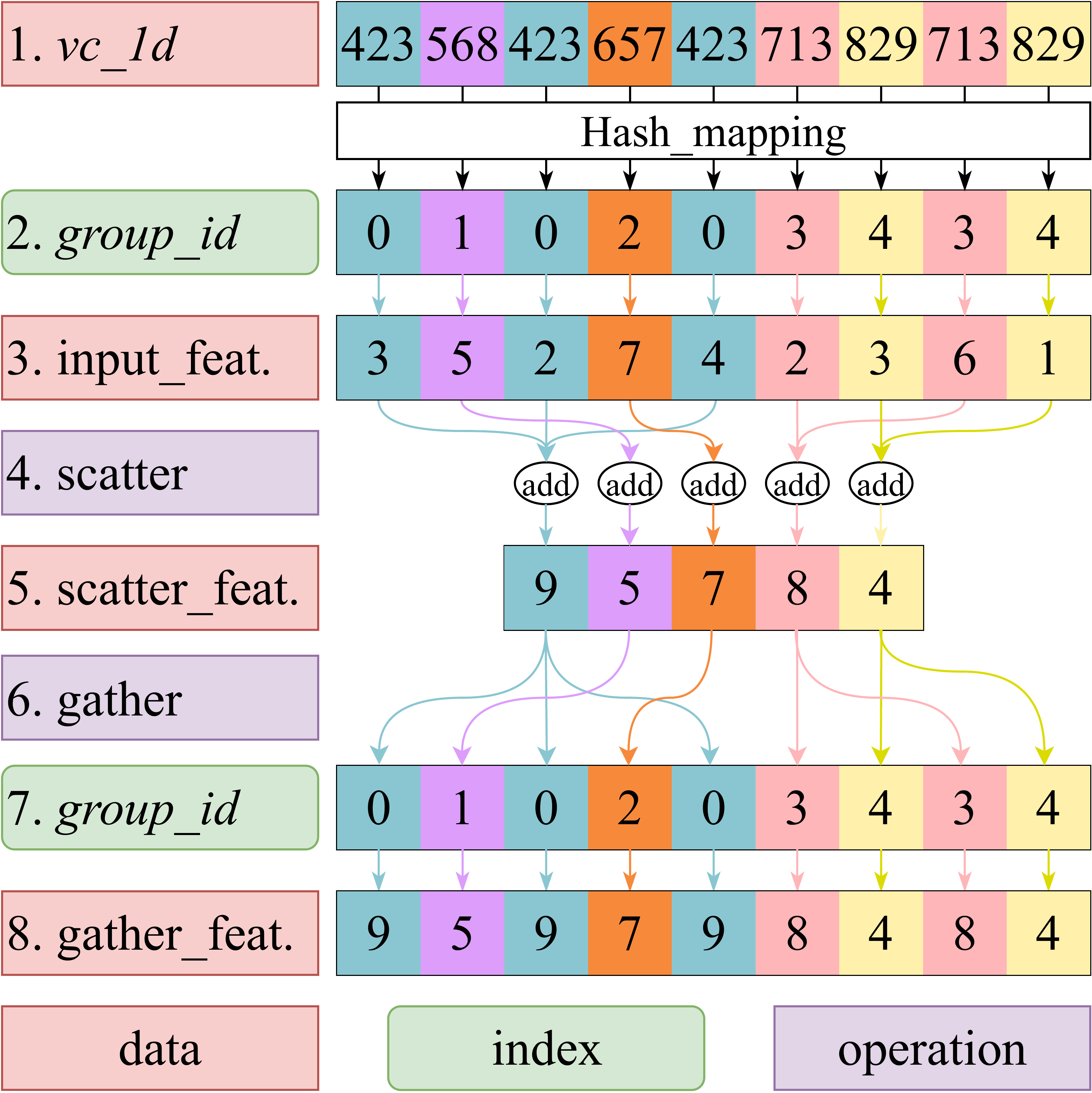}
        \caption{Scatter and gather operations}
        \label{fig:scatter}
        \end{center}
    \end{minipage}
    \vspace{-3mm}
\end{figure}

\begin{algorithm}[t]
\footnotesize
\caption{Intra-voxel Query}\label{alg1}
\SetAlgoLined
\SetKwInOut{Input}{Input}\SetKwInOut{Output}{Output}
\Input{Input Point Cloud \text{\bfseries\textit{$P_{in}[b,x,y,z]$}}}
\Output \text{\bfseries\textit{intra\_gid}}

\textbf{Initialization} \\
$voxel\_size$: voxel size used for sampling \\
$max\_r[x,y,z], min\_r[x,y,z]$: the upper and lower bound range of point cloud \\
$N[x,y,z]$ :  the number of voxels on x,y,z axis\\

\SetKwProg{myproc}{Procedure}{\hspace{0.1cm} $\textbf{Get\_3d\_Voxel\_Coords}(\mathbf{Points},voxel\_size)$}{Return\hspace{0.1cm}$\mathbf{V[b,x,y,z]}$}
\myproc{}{
    $\mathbf{V[b]} = \mathbf{Points[b]}$; \\
    $\mathbf{V[x,y,z]} = \lfloor (\mathbf{Points[x,y,z]} -min\_r[x,y,z])/voxel\_size\rfloor$; 
}

\SetKwProg{myproc}{Procedure}{\hspace{0.1cm} $\textbf{Flatten\_3d\_to\_1d}(\mathbf{V},voxel\_size)$}{Return\hspace{0.1cm}$\mathbf{coord\_1d}$}
\myproc{}{
    $N[x,y,z] =(max\_r[x,y,z] - min\_r[x,y,z])/voxel\_size$; \\
    $\mathbf{coord\_1d} = \mathbf{V[b]}\cdot N[x]\cdot N[y]\cdot N[z] + \mathbf{V[x]}\cdot N[y]\cdot N[z] + \mathbf{V[y]}\cdot N[z] + \mathbf{V[z]}$; 
}
\BlankLine

$\text{\bfseries\textit{vc\_3d}} = {Get\_3d\_Voxel\_Coords}(\text{\bfseries\textit{$P_{in}$}}, voxel\_size)$; \\
$\text{\bfseries\textit{vc\_1d}} = {Flatten\_3d\_to\_1d}({\text{\bfseries\textit{vc\_3d}}}, voxel\_size)$; \\
$\text{\bfseries\textit{intra\_gid}} = {hash\_func}(\text{\bfseries\textit{vc\_1d}})$; \\
\Return \text{\bfseries\textit{intra\_gid}}
\end{algorithm}

\textbf{Intra-voxel Query}. Its flow is shown in Alg.~\ref{alg1}. 
For input point set $P_{in}$, we first calculate the 3D voxel coordinates for each point and then flatten the 3D coordinates \textit{vc\_3d} into 1D coordinates \textit{vc\_1d}. The $\lfloor \cdot \rfloor$ is the floor function. If two points have the same \textit{vc\_1d}, it indicates that they are in the same voxel. Using \textit{vc\_1d} as the key, we obtain the $intra\_gid$ for each point through hash mapping. 
The $intra\_gid$ is short for $intra\_group\_{id}$.
If there are $M$ non-empty voxels in the current batch, then $intra\_gid \in \{0, 1, 2, \ldots, M-1\}$. As shown in Fig.~\ref{fig:vsa}(a), points with the same color within the same voxel are assigned to the same $intra\_gid$.
The larger the \textit{vc\_1d}, the larger the corresponding $intra\_gid$, as illustrated in the first two rows of Fig.~\ref{fig:scatter}. Assuming the values of \textit{vc\_1d} are $\{423, 568, 657, 713, 829\}$, their corresponding $intra\_gid$ values are $\{0, 1, 2, 3, 4\}$. 
As the $voxel\_size$ in Alg.~\ref{alg1} can take any value, it allows for downsampling with arbitrary sizes in multi-layer networks.

\textbf{Intra-voxel Aggregation}. 
Since the number of points in each voxel is usually unequal, $scatter$ operations are employed during feature aggregation. $Scatter$ is a CUDA kernel library capable of various parallel computations on matrices based on $group-index$, including summation, maximum, average, and softmax operations, as shown in rows $2$-$5$ of Fig.~\ref{fig:scatter}.
$Gather$ is the inverse operation of $scatter$, and by indexing and slicing the aggregated data. It aligns the aggregated data with the original input shape, as shown in rows $5$-$8$ of Fig.~\ref{fig:scatter}.
For multiple points within the same voxel, we aggregate their coordinates and features onto a single sampled point. We take the average of the coordinates of multiple points for downsampling:

\vspace{-0.3cm}
\begin{equation}
P_s = Mean_{scatter}(P_{in}, intra\_gid),
\end{equation}

where $P_s$ represents the downsampled point cloud.
After downsampling, there is only one sampled point within each voxel. 
For feature aggregation within voxels, local structural information is highly significant. We compute the local offset $\Delta P_{is}$ between multiple points within the voxel and the sampled point, concatenate this with $F_{in}$ (features of input points), and then apply MLP and max-pooling to obtain the aggregated features $F_s$.
\vspace{-0.2cm}
\begin{equation}
\Delta P_{is} = P_{in} - Gather(P_s, intra\_gid),
\label{eq:p_is}
\end{equation}

\vspace{-0.5cm}

\begin{equation}
F_s = Max_{scatter}(MLP(F_{in} \oplus  \Delta P_{is}) , intra\_gid).
\end{equation}

In Eq.~\ref{eq:p_is}, the $Gather$ operation aligns $M$ sampled points with the $N$ input points, as shown in rows $5$-$8$ of Fig.~\ref{fig:scatter}. In Fig.~\ref{fig:vsa}(b), each voxel contains only one sampled point, which aggregates the features of all points within the voxel.

\begin{figure}[t]
    \centering
    \includegraphics[width=0.48\textwidth]{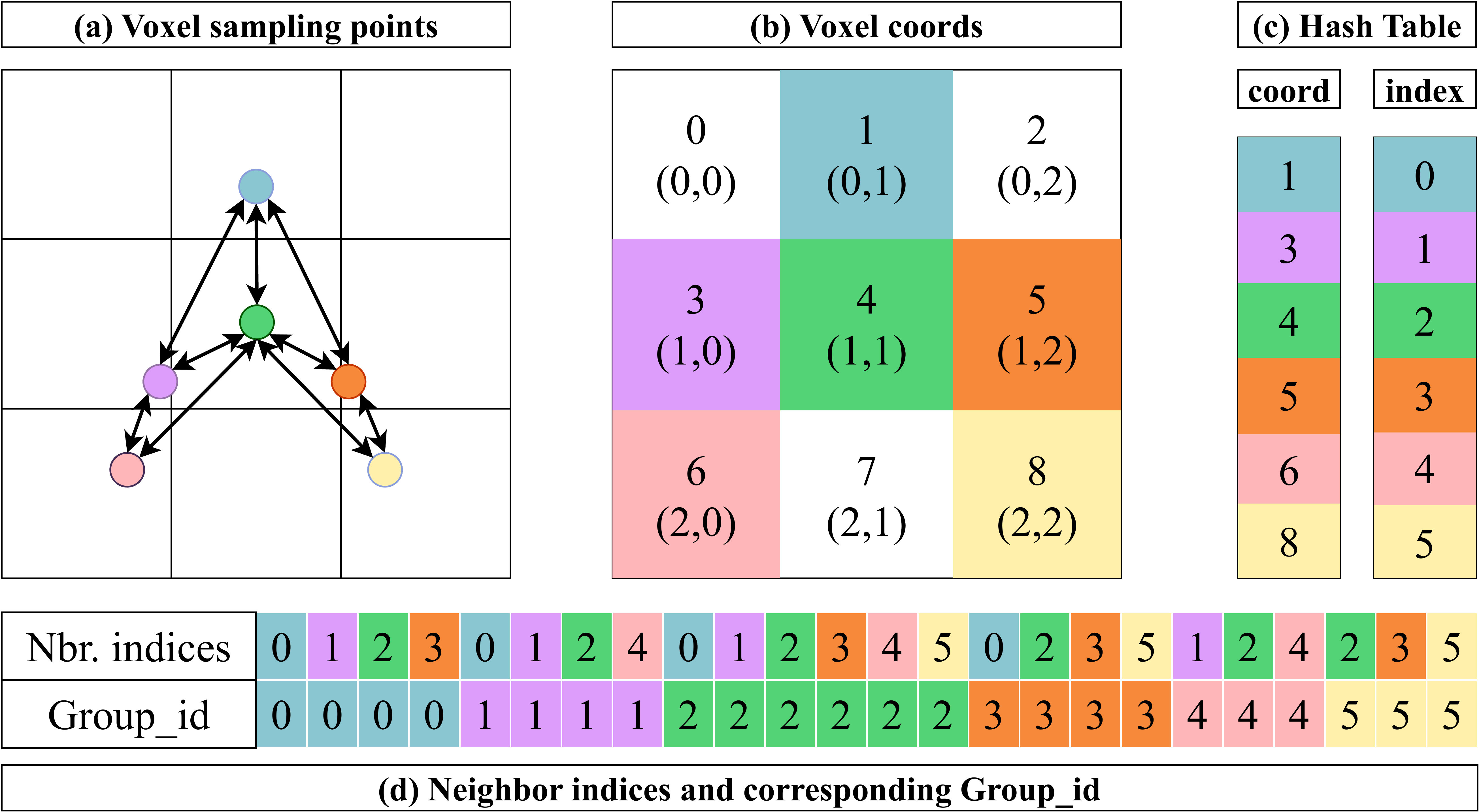}
    \caption{2D example of Inter-voxel Query. (a) The spatial distribution of sampled points. (b) (0,0)-(2,2) represents the 2D coordinates of voxels, 0-8 are flattened \textit{vc\_1d} values, and 0,2,7 are empty voxels. (c) Hash table between \textit{vc\_1d} and voxel\_index. (d) Neighbor indices and corresponding $group\_id$.}
    \label{fig:inter_query}
      \vspace{-5mm}
\end{figure}

\textbf{Inter-voxel Query}. The Intra-VFE step allows feature aggregation for point clouds at arbitrary voxel sizes. 
To further expand the receptive field, we perform feature extraction for neighboring points of the sampled points. Since the aggregated sampled points have a one-to-one correspondence with voxels, we can efficiently perform the neighbor search using the Inter-voxel Query. The process of the Inter-voxel Query is outlined in Alg.~\ref{alg2}, and a schematic diagram is given in Fig.~\ref{fig:inter_query}.

The algorithm consists of three steps:

1) Given voxel-sampled point $P_s$, we first calculate its corresponding \textit{vc\_1d}, then obtain the voxel indices ($V_{indices}$) in memory with a hash mapping. We use \textit{vc\_1d} as the key and $V_{indices}$ as the value to construct a hash table $T_{hash}$.

2) Generate the {local offset} $O_{nbr}$ for 3D voxels based on the hyperparameter $nbr\_size$. When $nbr\_size=3$, $O_{nbr}$ $= [[-1,-1,-1],[-1,-1,0], \ldots,[0,0,0] , \ldots,[1,1,1]]$, totaling 27 neighboring coordinate offsets.
For voxel coordinate $vc\_{3d}$ of voxel-sampled point, add the {local offset} $O_{nbr}$ to it, resulting in $[M,nbr\_size^3,3]$ neighborhood coordinate matrix ${nbr\_3d}$ using broadcasting. Construct $[0,1,2, \ldots, M-1]^T$ with each element repeated $nbr\_size^3$ times as $inter\_gid$.

3) Within the $nbr\_size^3$ neighborhood of each non-empty voxel, there may be some empty voxels. We remove the empty voxels with hash lookup. For the remaining non-empty voxel set $nbr\_ne$, we use voxel coordinate values as keys to find their corresponding neighborhood indices $nbr\_{indices}$ in the $T_{hash}$. Finally, return neighborhood indices and corresponding grouping information for each nonempty voxel.

\textbf{Inter-voxel Aggregation}. Utilizing $nbr\_{indices}$, we can get neighborhood feature matrices $F_{nbr}$ from non-empty voxel features $F_s$, as shown in Eq.~\ref{eq:f_nbr}. The relative position encoding $\Delta P_{inter}$ of sampled points and their neighbor points is obtained by Eq.~\ref{eq:p_inter}.
We can use various operators to extract neighborhood features. As shown in Eq.~\ref{eq:f_out}, $f$ represents the neighborhood feature extraction operator, which can be an MLP, multi-head attention~\cite{vaswani2017attention}, or vector attention~\cite{wu2022point}, etc. 
This paper uses vector attention~\cite{wu2022point} by default.

\vspace{-0.3cm}

\begin{equation}
F_{nbr} = Gather(F_s, nbr\_{indices}),
\label{eq:f_nbr}
\end{equation}

\vspace{-0.5cm}
\begin{equation}
P_{nbr} = Gather(P_s, nbr\_{indices}),\\
\end{equation}

\vspace{-0.5cm}

\begin{equation}
P_{center} = Gather(P_s, inter\_gid),\\
\end{equation}

\vspace{-0.5cm}

\begin{equation}
\Delta P_{inter} = P_{nbr} - P_{center},\\
\label{eq:p_inter}
\end{equation}

\vspace{-0.5cm}
\begin{equation}
\label{eq:f_out}
F_{out} = Sum_{scatter}(f(F_{nbr},\Delta P_{inter}), inter\_gid).
\end{equation}
\vspace{-0.4cm}

\begin{algorithm}[t]
\footnotesize
   \caption{Inter-voxel Query}\label{alg2}
   \SetAlgoLined
   \SetKwInOut{Input}{Input}\SetKwInOut{Output}{Output}
   \Input{Voxel sampled points \text{\bfseries\textit{$P_s[B,x,y,z]$}}}
   \Output{\text{\bfseries\textit{nbr\_indices, inter\_gid}}}

   \textbf{Initilization} \\
   $voxel\_size$: voxel size used for partitioning \\
   $\hspace{0.3cm}nbr\_size$: the number of neighbor points in each axis \\
   
    \SetKwProg{myproc}{Procedure}{\hspace{0.1cm} $\textbf{Generate\_Local\_Offset}(nbr\_size)$}{Return\hspace{0.1cm}\textit{Local\_Offset}}
    \myproc{}{
        \textit{Local\_Offset}$ \leftarrow [\hspace{0.1cm}];$\\
        $c = \lfloor nbr\_size / 2 \rfloor$; \\
        \For{$x$ $\mathbf{in}$ $range(nbr\_size)$}{
        \For{$y$ $\mathbf{in}$ $range(nbr\_size)$}{
        \For{$z$ $\mathbf{in}$ $range(nbr\_size)$}{
        \textit{Local\_Offset}$.append(\lbrack x-c, y-c, z-c \rbrack)$;
        }
        }
        }
    }
    \BlankLine
    
    $\text{\bfseries\textit{vc\_3d}} = {Get\_3d\_Voxel\_Coords}(\text{\bfseries\textit{$P_s$}}, voxel\_size)$; \\
    $\text{\bfseries\textit{vc\_1d}} = {Flatten\_3d\_to\_1d}(\text{\bfseries\textit{vc\_3d}}, voxel\_size)$; \\
    $\text{\bfseries\textit{V}}_\text{\bfseries\textit{indices}} = {hash\_func}(\text{\bfseries\textit{vc\_1d}})$; \\
    $\text{\bfseries\textit{T}}_\text{\bfseries\textit{hash}} = \lbrack \text{\bfseries\textit{vc\_1d}},\text{\bfseries\textit{V}}_\text{\bfseries\textit{indices}} \rbrack$; \\
    ${O_{nbr}} = \textit{Generate\_Local\_Offset}(nbr\_size) $; \\
    $\text{\bfseries\textit{nbr\_3d}} = \text{\bfseries\textit{vc\_3d[x,y,z]}} + {O_{nbr}}; $\\
    $\text{\bfseries\textit{inter\_gid}} = [0,1,2,\ldots,len(\text{\bfseries\textit{vc\_3d}})-1] $; \\
    $\text{\bfseries\textit{inter\_gid}} = \text{\bfseries\textit{inter\_gid}}.repeat(1,nbr\_size^3)$; \\
    $\text{\bfseries\textit{nbr\_1d}} = {Flatten\_3d\_to\_1d}(\text{\bfseries\textit{nbr\_3d}}, voxel\_size) $; \\
    $\text{\bfseries\textit{mask}} = {Find\_non\_empty\_voxels}(\text{\bfseries\textit{nbr\_1d}}, \text{\bfseries\textit{vc\_1d}}) $; \\
    $\text{\bfseries\textit{nbr\_ne}} = \text{\bfseries\textit{nbr\_1d}} \lbrack \text{\bfseries\textit{mask}} \rbrack $; \\
    $\text{\bfseries\textit{inter\_gid}} = \text{\bfseries\textit{inter\_gid}} \lbrack \text{\bfseries\textit{mask}} \rbrack $; \\
    $\text{\bfseries\textit{nbr\_indices}} = {hash\_find}(\text{\bfseries\textit{nbr\_ne}}, \text{\bfseries\textit{T}}_\text{\bfseries\textit{hash}})$; \\
   \Return{\text{\bfseries\textit{nbr\_indices}}, \text{\bfseries\textit{inter\_gid}}}
   \BlankLine
\end{algorithm}

\textbf{Complexity comparison.}
We denote $N$ as the number of input points, $M$ as the number of sampled points, and $K$ as the number of neighbors to search. Both $M$ and $N$ are much larger than $K$ in a large-scale point cloud. 
In previous point-based methods~\cite{qi2017pointnet++,zhao2021point,qian2022pointnext,zhang2023starting}, the complexities of FPS and KNN are $\mathcal{O}(MN)$ and $\mathcal{O}(MNlogK)$, respectively. In our method, the complexities of Intra-voxel Query and Inter-voxel Query are $\mathcal{O}(N)$ and $\mathcal{O}(MK)$. In this paper, $K = nbr\_size^3$, where $nbr\_size$ is typically equal to $3$. Our approach shares similar complexity with other voxel-based methods~\cite{graham20183d,choy20194d,he2022voxel}. The key distinction is that our method retains precise coordinates of sampled points and allows for sampling and neighbor search at arbitrary voxel sizes at each layer.

\subsection{Voxel Adaptation Module} \label{Sec:VAM}

Voxel size has a significant impact on the performance of voxel-based methods. The most crucial parameter affecting performance is the number of voxel sampled points. FPS can precisely control the downsampling ratio by controlling the number of sampled points. However, the number of sampled points is unknown and dynamic for each point cloud frame when using voxel sampling. We propose using a fixed downsampling ratio as a reference to automatically adjust the voxel size, ensuring that the average sampling ratio over the entire dataset approaches this reference ratio. The proposed Voxel Adaptation Module (VAM) is used as a preprocessing module. Before model training, we use VAM to determine the voxel size for each downsampling layer. The voxel size remains constant during the training process.

In 3D space, the total number of voxels $N_{voxels}$, including empty voxels, is given by:

\vspace{-0.5cm}
\begin{equation}
\centering
N_{voxels} = \frac{V_{space}}{voxel\_size^3},  
\label{eq:N_voxels}
\end{equation}
where $V_{space}$ is the volume of the 3D space.
Since the distribution of point clouds in space is non-uniform, there is no strict equality relationship in Eq.~\ref{eq:N_voxels}. Intuitively, for the point clouds of the entire dataset, it is generally true that the larger the voxel size, the fewer the total number of non-empty voxels. This relationship can be approximately expressed as:

\begin{equation}
\label{eq:vsize_and_ratio1}
\centering
N_{s} \propto\ \frac{N_{i}}{voxel\_size^3},
\end{equation}
\vspace{-0.2cm}
\begin{equation}
\label{eq:vsize_and_ratio2}
\centering
voxel\_size \propto\ \sqrt[3]{\frac{N_{i}}{N_{s}}},
\end{equation}
where $N_{i}$ and $N_{s}$ represent the sum of the number of input points and the sum of the number of voxel sampled points across the entire dataset.
To make the sampling voxel size variable, we introduce a variable factor $Scale$. Therefore, the varable voxel size can be formulated as $voxel\_size= V_0 \cdot \text{e}^{{Scale}}$, where $V_0$ is a constant value of the initial voxel size.

\begin{figure}[t]
    \begin{center}
          \vspace{-0.1cm}
        \includegraphics[width=\linewidth]{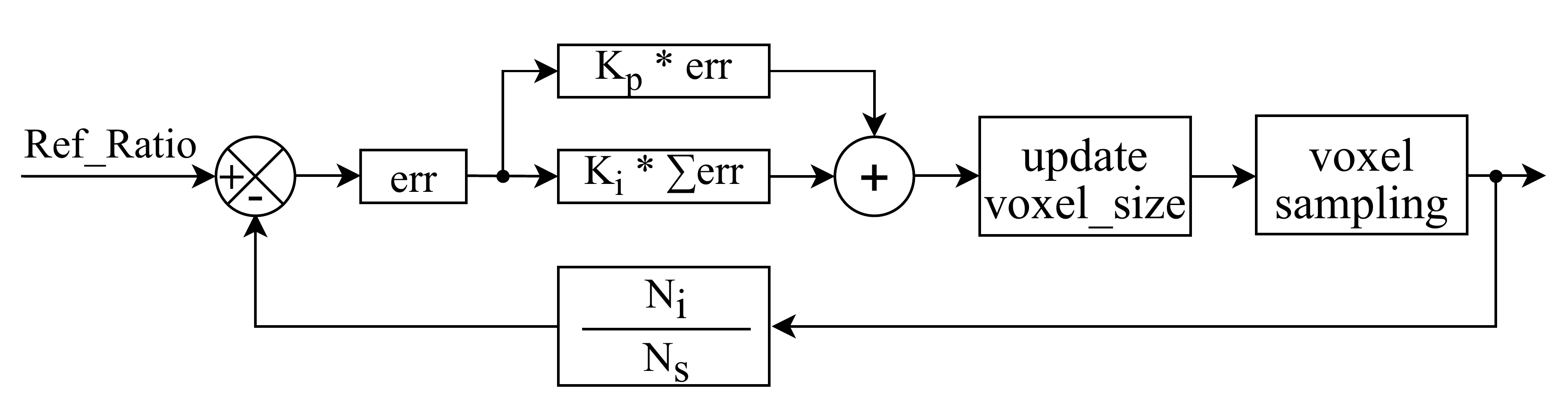}
    \end{center}
   \caption{Voxel Adaptation Module via the PI control algorithm}
   \label{fig5_voxel_adaptation}
\end{figure}

Since the mapping of input points to sampled points is obtained through a scatter operation on point indices, there is no gradient flow between the number of points and $voxel\_size$, making it impossible to train directly via the network. Inspired by the Proportional-Integral (PI) algorithm in automatic control, we have designed an algorithm for automatic voxel size adjustment. The algorithm's flow and schematic are shown in Alg.~\ref{alg3} and Fig.~\ref{fig5_voxel_adaptation}.

\begin{algorithm}[t]
\footnotesize
   \caption{VAM flow}\label{alg3}
   \SetAlgoLined
\Do{$err > \epsilon$}{
    $\text{\romannumeral 1.}\ err = Ref\_Ratio - {N_{i}}\text{ / }{N_{s}}$\; \\
    $\text{\romannumeral 2.}\ \textit{Diff} = K_p \cdot err + K_i \cdot \sum err$\; \\
    $\text{\romannumeral 3.}\ Scale = Scale + I_r \cdot (\text{sigmoid}(\textit{Diff}) - 0.5)$\; \\
    $\text{\romannumeral 4.}\ voxel\_size = V_0 \cdot \text{e}^{Scale}$\; \\
}
\end{algorithm}

We next introduce VAM in detail. Alg.~\ref{alg3}-\text{\romannumeral 1}: For voxel centroid sampling, given a downsampling ratio \textit{Ref\_Ratio}, we calculate the deviation $err$ by subtracting the actual sampling ratio corresponding to the current voxel size across the entire dataset from \textit{Ref\_Ratio}. Alg.~\ref{alg3}-\text{\romannumeral 2}: Integrate $err$ to obtain the cumulative deviation $\sum{err}$, thus further obtaining the output \textit{Diff} of the PI algorithm. Alg.~\ref{alg3}-\text{\romannumeral 3}: We adjust the scaling factor $Scale$ based on \textit{Diff}. When \textit{Diff} $> 0$, it indicates that $N_{s}$ is too large, and the current voxel size is too small, so the $Scale$ increases. When \textit{Diff} $< 0$, the $Scale$ decreases. Alg.~\ref{alg3}-\text{\romannumeral 4}: We calculate the actual voxel size that participates in the sampling based on $Scale$. Voxel\_size is a monotonically increasing function with respect to $Scale$ and ensures non-negativity. The initial value of the $Scale$ is set to 0. We iterate over the entire dataset using Alg.~\ref{alg3}-(\text{\romannumeral 1}-\text{\romannumeral 4}). When $err$ is less than an extremely small value $\epsilon$, we consider it as reaching the convergence condition. At this point, $voxel\_size$ ensures that the average sampling ratio across the entire dataset approximates the given value \textit{Ref\_Ratio}. In Alg.~\ref{alg3}, $V_0$, $I_r$, $K_p$, $K_i$, and $\epsilon$ are hyperparameters. $V_0$ is the initial voxel size, and $I_r$ is a parameter similar to the learning rate. $K_p$ and $K_i$ control the ratio of ${err}$ and $\sum{err}$, and $\epsilon$ controls the convergence condition. $\epsilon$ usually set to $0.1\%$.

\section{Experiments}
\label{sec:Experiments}

\begin{table*}
\centering
\footnotesize
    \caption{Quantitative detection performance achieved by different methods on the Waymo {validation} set (100\% training data, single-frame setting). The best results are shown in bold.} 
   \resizebox{0.9\textwidth}{!}{
   \begin{tabular}{c c cc cc cc cc cc cc}
\toprule
   \multirow{2}{*}{Method} & \multirow{2}{*}{Reference} & \multicolumn{2}{c}{Vehicle (L1)} &\multicolumn{2}{c}{Vehicle (L2)} & \multicolumn{2}{c}{Pedestrian (L1)} & \multicolumn{2}{c}{Pedestrian (L2)} & \multicolumn{2}{c}{Cyclist (L1)} & \multicolumn{2}{c}{Cyclist (L2)} \\
  \cmidrule{3-14}
   & & mAP & mAPH & mAP & mAPH & mAP & mAPH & mAP & mAPH & mAP & mAPH & mAP & mAPH \\
   \midrule
    SECOND \cite{yan2018second}                 &Sensors'18  & 72.3 & 71.7 & 63.9 & 63.3 & 68.7 & 58.2 & 60.7 & 51.3 & 60.6 & 59.3 & 58.3 & 57.0 \\
    PointPillars \cite{lang2019pointpillars}    &CVPR'19     & 72.1 & 71.5 & 63.6 & 63.1 & 70.6 & 56.7 & 62.8 & 50.3 & 64.4 & 62.3 & 61.9 & 59.9 \\
    CenterPoint-Pillar \cite{yin2021center}      &CVPR'21     & 73.4 & 72.9 & 65.1 & 64.6 & 75.4 & 65.1 & 67.6 & 58.3 & 67.8 & 66.2 & 65.3 & 63.8  \\
    Part-$A^2$ \cite{shi2020points}             &TPAMI'20    & 77.1 & 76.5 & 68.5 & 68.0 & 75.2 & 66.9 & 66.2 & 58.6 & 68.6 & 67.4 & 66.1 & 64.9 \\
    PV-RCNN \cite{shi2020pv}                    &CVPR'20     & 78.0 & 77.5 & 69.4 & 69.0 & 79.2 & 73.0 & 70.4 & 64.7 & 71.5 & 70.3 & 69.0 & 67.8 \\
    PV-RCNN++ \cite{shi2022pv}                  &IJCV'22     & 79.3 & 78.8 & 70.6 & 70.2 & 81.8 & 76.3 & 73.2 & 68.0 & 73.7 & 72.7 & 71.2 & 70.2 \\
    SST~\cite{fan2021embracing}                 &CVPR'22     & 74.2 & 73.8 & 65.5 & 65.1 & 78.7 & 69.6 & 70.0 & 61.7 & 70.7 & 69.6 & 68.0 & 66.9 \\
    VoxSeT~\cite{voxelset}                      &CVPR'22     & 74.5 & 74.0 & 66.0 & 65.6 & 80.0 & 72.4 & 72.5 & 65.4 & 71.6 & 70.3 & 69.0 & 67.7 \\
    PillarNet-34~\cite{shi2022pillarnet}        &ECCV'22     & 79.1 & 78.6 & 70.9 & 70.5 & 80.6 & 74.0 & 72.3 & 66.2 & 72.3 & 71.2 & 69.7 & 68.7 \\
    CenterFormer~\cite{Zhou_centerformer}       &ECCV'22     & 75.0 & 74.4 & 69.9 & 69.4 & 78.6 & 73.0 & 73.6 & 68.3 & 72.3 & 71.3 & 69.8 & 68.8 \\
    FlatFormer~\cite{liu2023flatformer}         &CVPR'23     & 77.5 & 77.1 & 69.0 & 68.6 & 79.6 & 73.0 & 71.5 & 65.3 & 71.3 & 70.1 & 68.6 & 67.5 \\
    DSVT-Voxel~\cite{DSVT}                      &CVPR'23     & 79.7 & 79.3 & 71.4 & 71.0 & 83.7 & 78.9 & 76.1 & 71.5 & {77.5} & {76.5} & {74.6} & {73.7} \\
    VoxelNeXt~\cite{chen2023voxelnext}          &CVPR'23     & 78.2 & 77.7 & 69.9 & 69.4 & 81.5 & 76.3 & 73.5 & 68.6 & 76.1 & 74.9 & 73.3 & 72.2 \\
    PA-RCNN~\cite{lu2023improving}              &TIV'23      & 78.8 & 78.3 & 70.4 & 70.0 & 81.7 & 76.5 & 73.5 & 68.6 & 74.2 & 73.2 & 71.5 & 70.5 \\
    KPDet~\cite{kpdet}                          &Neucom.'24      & 79.5 & 79.0 & 70.4 & 70.0 & 84.0 & 77.7 & 75.4 & 69.5 & 75.6 & 74.6 & 73.6 & 72.6 \\
    \rowcolor{cyan! 15}AVS-Net (ours)           & -         & \textbf{79.9} & \textbf{79.4} & \textbf{71.7} &	\textbf{71.3} & {84.0} & \textbf{79.3} & \textbf{76.4} &	\textbf{72.0} & \textbf{77.8} & \textbf{76.7} & \textbf{75.0} & \textbf{73.9} \\ 
    \bottomrule
   \end{tabular}
   }
   \vspace{-0.3cm}
   \label{tab:Waymo_val}
\end{table*}

\subsection{3D Object Detection}
Waymo Open Dataset~\cite{sun2020scalability} is one of the largest and richest autonomous driving datasets. 
We validate the effectiveness of AVS-Net on the 3D object detection task. The experimental results on the Waymo~\cite{sun2020scalability} validation set are shown in Table~\ref{tab:Waymo_val}. Specifically, 
we choose the framework of CenterPoint-Pillar~\cite{yin2021center} and use four-layer VSA modules as 3D backbone network before BEV network.
The detection accuracy for vehicles, pedestrians, and cyclists significantly improved after using our AVS-Net. 
Compared to the prior best method DSVT-Voxel~\cite{DSVT}, our AVS-Net also achieves better results in all categories.

\begin{figure}
   \begin{center}
      \includegraphics[width=0.45\textwidth]{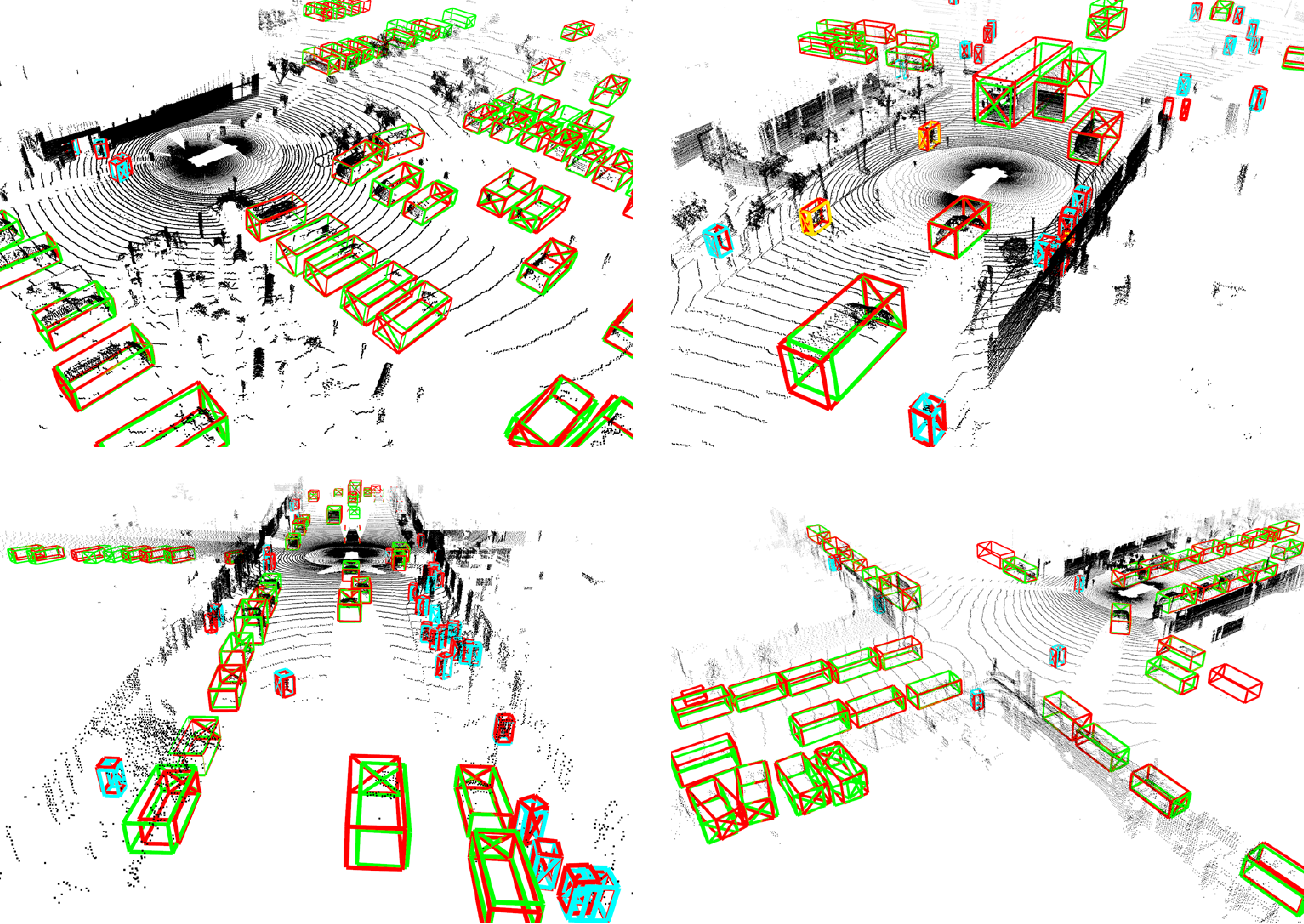}
   \end{center}
   \vspace{-0.5cm}
   \caption{Qualitative results achieved by our AVS-Net on the Waymo validation set. Note that, the ground-truth bounding boxes are shown in red, and the predicted bounding boxes are shown in green for vehicle, cyan for pedestrian, and yellow for cyclist. Best viewed in color.}
   \label{fig:waymo_vis}
   \vspace{-0.3cm}
\end{figure}

Tablees ~\ref{tab:waymo-dist-car}, ~\ref{tab:waymo-dist-ped}, and ~\ref{tab:waymo-dist-cyc} present the 3D detection results of three classes of objects divided by distance. Except for BtcDet with the generation module showing advantages in vehicle detection within 0-30m, AVS-Net outperforms other methods in all other metrics. This is attributed to AVS-Net effectively preserving the fine geometric structures of foreground objects during downsampling, thus achieving better detection results for distant and small objects. We also demonstrate the qualitative results obtained by our AVS-Net in Fig. ~\ref{fig:waymo_vis}. As observed from Fig. ~\ref{fig:waymo_vis}, the proposed AVS-Net is capable of detecting dense vehicles and pedestrians from complex scenes, exhibiting good detection performance even for distant objects.

\begin{table}[t]
    \caption{Performance comparison on the Waymo {validation} set for 3D vehicle (IoU~=~0.7) detection by range.}
    \centering
    
    \resizebox{0.45\textwidth}{!}{
        \begin{tabular}{ccccc}
            \hline
            \multirow{2}{*}{Method} &
            \multirow{2}{*}{Reference} &
            \multicolumn{3}{c}{Vehicle 3D mAP/mAPH}    \\
            
            \cline{3-5}
            
            {}   & & 0-30m  & 30-50m & 50m-inf    \\ \hline
            \textit{\textbf{LEVEL\_1 (IoU=0.7):}}       & &&& \\
            SECOND \cite{yan2018second}                              &Sensors'18           &90.66/- &70.03/- &47.55/-   \\
            PointPillars \cite{lang2019pointpillars}                 &CVPR'19           &81.01/- &51.75/- &27.94/-   \\
            Part-$A^2$ \cite{shi2020points}                          &TPAMI'20           &92.35/- &75.91/- &54.06/-  \\
            Voxel-RCNN \cite{deng2021voxel}                          &AAAI'21           &92.49/- &74.09/- &53.15/-   \\
            CT3D \cite{sheng2021improving}                           &ICCV'21           &92.51/- &75.07/- &55.36/-  \\
            PV-RCNN \cite{shi2020pv}                                 &CVPR'20           &92.96/- &76.47/- &55.96/-  \\
            PV-RCNN++ \cite{shi2022pv}                               &IJCV'22           &93.34/- &78.08/- &57.19/-  \\
            VoxSeT \cite{voxelset}                                   &CVPR'22           &92.78/- &77.21/- &54.41  \\
            PDV \cite{hu2022point}                                   &CVPR'22           &93.13/92.71 &75.49/74.91 &54.75/53.90  \\
            BtcDet \cite{xu2022behind}                               &AAAI'22           &\textbf{96.11}/- &77.64/- &54.45/-  \\
            PA-RCNN \cite{lu2023improving}                           &TIV'23            &92.88/92.48 &77.60/77.10 &57.71/56.99  \\
            KPDet~\cite{kpdet}                                       &Neucom.'24         &93.00/-  & 77.68/-  & 57.54/-   \\
            \rowcolor{cyan! 15}AVS-Net (ours)                        &-                 &93.84/\textbf{93.47} &\textbf{78.46/77.97} &\textbf{58.47/57.82}  
            \\ \hline \hline
            \textit{\textbf{LEVEL\_2 (IoU=0.7):}}       & &&& \\
            PV-RCNN \cite{shi2020pv}                                 &CVPR'20     &91.58/- &65.13/- &36.46/-  \\
            Voxel-RCNN \cite{deng2021voxel}                          &AAAI'21     &91.74/- &67.89/- &40.80/-  \\
            CT3D \cite{sheng2021improving}                           &ICCV'21     &91.76/- &68.93/- &42.60/-  \\
            VoxSeT \cite{voxelset}                                   &CVPR'22     &92.05/- &70.10/- &43.20/-  \\
            PDV \cite{hu2022point}                                   &CVPR'22     &92.41/91.99 &69.36/68.81 &42.16/41.48  \\
            BtcDet \cite{xu2022behind}                               &AAAI'22     &\textbf{95.99}/- &70.56/- &43.87/-  \\
            PA-RCNN \cite{lu2023improving}                           &TIV'23     &91.69/90.91 &71.20/70.73 &45.30/44.71  \\
            \rowcolor{cyan! 15}AVS-Net (ours)                        &-          &92.56/\textbf{92.19} &\textbf{72.04/71.59} &\textbf{45.94/45.40}  
            \\ \hline
            
        \end{tabular}
    }
    \label{tab:waymo-dist-car}
\end{table}

\begin{table}
\centering
\caption{Performance comparison on the Waymo {validation} set for 3D pedestrian (IoU~=~0.5) detection by range.}
\resizebox{0.45\textwidth}{!}{
\begin{tabular}{ccccc}
\hline
        \multirow{2}{*}{Method} 
        &   \multirow{2}{*}{Reference} 
        & \multicolumn{3}{c}{Pedestrian 3D mAP/mAPH} \\
                        \cline{3-5}
                        && 0-30m        & 30-50m       & 50m-Inf     \\ \hline
                        
            \textit{\textbf{LEVEL\_1 (IoU=0.5):}}       & &&& \\
            SECOND \cite{yan2018second}                          &Sensors'18                &74.39/- &67.24/- &56.71/-   \\
            PointPillars \cite{lang2019pointpillars}             &CVPR'19               &67.99/- &57.01/- &41.29/-   \\
            Part-$A^2$ \cite{shi2020points}                      &TPAMI'20               &81.87/- &73.65/- &62.34/-  \\
            PV-RCNN \cite{shi2020pv}                             &CVPR'20               &83.33/- &78.53/- &69.36/-  \\
            PV-RCNN++ \cite{shi2022pv}                           &IJCV'22                &84.88/- &79.65/- &70.64/-  \\
            PDV \cite{hu2022point}                              &CVPR'22                &80.32/73.60  & 72.97/63.28  & 61.69/50.07  \\
            PA-RCNN \cite{lu2023improving}                      &TIV'23                &86.13/82.17 &80.73/74.89 &73.14/64.57  \\
            KPDet~\cite{kpdet}                                  &Neucom.'24          &85.41/- & 79.51/- & 72.70/-   \\
            \rowcolor{cyan! 15}AVS-Net (ours)                   &-                  &\textbf{88.75/85.02} &\textbf{82.50/77.43} &\textbf{74.53/67.13}  
                        \\ \hline \hline
            \textit{\textbf{LEVEL\_2 (IoU=0.5):}}       & &&& \\
            PDV \cite{hu2022point}                           &CVPR'22             &75.26/68.82  & 65.78/56.85  & 47.46/38.30  \\
            PA-RCNN \cite{lu2023improving}                   &TIV'23             &81.41/77.55 &73.62/68.11 &59.04/51.70  \\
            \rowcolor{cyan! 15}AVS-Net (ours)                &-                                &\textbf{85.37/81.62} &\textbf{75.74/70.93} &\textbf{60.92/54.43}  
            \\ \hline
\end{tabular}
}
\label{tab:waymo-dist-ped}
\end{table}

\begin{table}
\centering
\caption{Performance comparison on the Waymo {validation} set for 3D cyclist (IoU~=~0.5) detection by range.}
\resizebox{0.45\textwidth}{!}{
\begin{tabular}{ccccc}
\hline
\multirow{2}{*}{Method} &   \multirow{2}{*}{Reference} & \multicolumn{3}{c}{Cyclist 3D mAP/mAPH} \\
                        \cline{3-5}
                        && 0-30m        & 30-50m       & 50m-Inf     \\ \hline
                        
            \textit{\textbf{LEVEL\_1 (IoU=0.5):}}       & &&& \\
            SECOND \cite{yan2018second}                            &Sensors'18             &73.33/- &55.51/- &41.98/-   \\
            Part-$A^2$ \cite{shi2020points}                        &TPAMI'20             &80.87/- &62.57/- &71.00/-  \\
            PV-RCNN \cite{shi2020pv}                             &CVPR'20               &81.10/- &65.65/- &52.58/-  \\
            PV-RCNN++ \cite{shi2022pv}                           &IJCV'22               &83.65/- &68.90/- &51.41/-  \\
            PDV \cite{hu2022point}                               &CVPR'22                &80.86/79.83  & 62.61/61.45  & 46.23/44.12  \\
            PA-RCNN \cite{lu2023improving}                       &TIV'23               &83.56/82.47 &70.84/69.68 &54.40/52.91  \\
            \rowcolor{cyan! 15}AVS-Net (ours)                    &-                &\textbf{86.25/85.18} &\textbf{74.83/73.63} &\textbf{60.14/59.10}  
                        \\ \hline \hline
            \textit{\textbf{LEVEL\_2 (IoU=0.5):}}       & &&& \\
            PDV \cite{hu2022point}                            &CVPR'22            &80.42/79.40  & 58.95/57.87  & 43.05/41.09  \\
            PA-RCNN \cite{lu2023improving}                    &TIV'23            &82.96/81.88 &67.03/65.93 &50.71/49.30  \\ 
            \rowcolor{cyan! 15}AVS-Net (ours)                 &-             &\textbf{85.63/84.57} &\textbf{70.76/69.62} &\textbf{56.10/55.13}  
            \\ \hline
\end{tabular}
}
\label{tab:waymo-dist-cyc}
\end{table}

\subsection{3D Semantic Segmentation}

ScanNet~\cite{dai2017scannet} is a large-scale dataset widely used for intelligent robot navigation. 
We follow the setup of PTv2~\cite{wu2022point} for training and evaluation. 
Table~\ref{tab:ScanNet_miou} presents the semantic segmentation results of our method compared to previous methods on ScanNet~\cite{dai2017scannet}.
In Table~\ref{tab:ScanNet_latency}, we compare the number of parameters, latency, and mIoU with recent methods. We reproduce the results of each method on the validation set using their official code. To ensure a fair comparison, all methods pre-sample point clouds with $2cm$ voxel size, and mIoU do not involve TTA (test-time augmentation). We use the same machine with an AMD EPYC 7313 CPU and a single NVIDIA 4090 GPU to measure the latency of methods.
In Table~\ref{tab:ScanNet_latency}, latency is the average time taken for inference on each frame's point cloud in the validation set. The batch size is set to 1.

In terms of accuracy, our AVS-Net achieves the best mIoU, as shown in Table~\ref{tab:ScanNet_latency}. 
In terms of efficiency, PointNeXt~\cite{qian2022pointnext} has the highest latency due to the FPS downsampling method. PTv2~\cite{wu2022point} employs efficient voxel sampling, which still incurs higher latency due to expensive KNN neighbor search. Our AVS-Net has similar latency to LargeKernel3D~\cite{chen2023largekernel3d} but achieves gains of $2.5\%$ of mIoU. Our AVS-Net attains the highest mIoU on large-scale point clouds while maintaining high operational efficiency.

\begin{table}
    \centering
    \caption{Semantic segmentation on ScanNet validation set. TTA is test time augmentaion.}
    \resizebox{0.95\linewidth}{!}{
        \begin{tabular}{l  c c c c }
    \toprule Method  &Reference  & Input &TTA & mIoU (\%)                \\
    \specialrule{0em}{2pt}{0pt}
    \hline
    \specialrule{0em}{2pt}{0pt}
    PointNet++~\cite{qi2017pointnet++}             & NeurIPS'17 & point & 
\ding{56} & 53.5                 \\
    PointConv~\cite{wu2019pointconv}               & CVPR'19 & point & 
\ding{56} & 61.0                \\
    PointASNL~\cite{yan2020pointasnl}              & CVPR'20 & point & 
\ding{56} & 63.5                 \\
    KPConv~\cite{thomas2019kpconv}                 & ICCV'19 & point & 
\ding{56} & 69.2                 \\
    PTv1~\cite{zhao2021point}                      & ICCV'21 & point & 
\ding{52} & 70.6                  \\
    PTv2 ~\cite{wu2022point}                       & NeurIPS'22 & point & 
\ding{52} & 75.4           \\
    PointNeXt~\cite{qian2022pointnext}             & NeurIPS'22 & point & 
\ding{52} & 71.5                \\
    ADS ~\cite{hong2023attention}                  & ICCV'23 & point & 
\ding{52} & 75.6 \\
    IEFM. ~\cite{iefm}                              & Neuro.'24 & point & 
\ding{52} & 73.9 \\
    \hline
    SparseConvNet~\cite{graham20183d}              & CVPR'18 & voxel & 
\ding{56} & 69.3                \\
    MinkowskiNet~\cite{choy20194d}                 & CVPR'19 & voxel & 
\ding{56} & 72.2                 \\
    FastPointTransformer~\cite{park2022fast}       & CVPR'22 & voxel & 
\ding{56} & 72.1                \\
    Stratifiedformer~\cite{lai2022stratified}& CVPR'22 & voxel & 
\ding{52} & 74.3                 \\
    LargeKernel3D ~\cite{chen2023largekernel3d}    & CVPR'23 & voxel & 
\ding{56} & 73.5                \\
    OctFormer ~\cite{Wang2023OctFormer}            & TOG'23  & voxel & \ding{52} & 75.7           \\
    
    \rowcolor{cyan! 20}AVS-Net \small{(ours)}                      & - & voxel & 
\ding{56} & \textbf{76.0}  \\
    \bottomrule
\end{tabular}

    }
    \label{tab:ScanNet_miou}
    \vspace{-2mm}
\end{table}

\subsection{Object Part Segmentation}
ShapeNetPart~\cite{yi2016scalable} is an important dataset for fine-grained 3D shape segmentation. 
We follow the setup in PointNeXt~\cite{qian2022pointnext} for training and evaluation. 
Regarding evaluation metrics, we report category mIoU and instance mIoU in Table~\ref{tab:shapenetpartresult}.
Although the performance in ShapeNetPart is quite saturated, our AVS-Net achieves the highest category mIoU and instance mIoU with considerable improvements ($+0.3\%$ category mIoU). 
The integer multiples expansion of voxel size has resulted in an unreasonable sampling ratio, making it difficult for previous models with voxel sampling to achieve good results on small objects. However, our method surpasses all previous methods in Table~\ref{tab:shapenetpartresult}, demonstrating that our AVS-Net is also effective on small-scale (few number of points) datasets.

\begin{table}[t]
    \begin{minipage}{\linewidth}
        \caption{Latency and mIoU without TTA on the ScanNet validation set.}
        \begin{center}
        \resizebox{0.9\linewidth}{!}{
        
\begin{tabular}{c c c c }
    \toprule Method      & Params. (M) & Latency (ms)           & mIoU (\%)       \\
    \hline
    PointNeXt-XL~\cite{qian2022pointnext}      & 41.6                 & 951.5        & 71.5      \\
    MinkowskiNet-34~\cite{choy20194d}          & 37.9                 & {87.8}         & 72.2      \\
    PTv2~\cite{wu2022point}                    & {11.3}                 & 181.4        & 74.7      \\
    LargeKernel3D~\cite{chen2023largekernel3d} & 40.2                 & 118.2          & 73.5      \\
    OctFormer~\cite{Wang2023OctFormer}         & 39.0                 & 109.5        & 74.5      \\
    
    \rowcolor{cyan! 20}AVS-Net \small{(ours)}                      & 20.9 & {120.4} & \textbf{76.0} \\
    \bottomrule
\end{tabular}

        }
        \label{tab:ScanNet_latency}
        \end{center}
    \end{minipage}%

    \vfill 
    \begin{minipage}{\linewidth}
    \caption{Part segmentation results on the ShapeNetPart validation set.}
    \begin{center}
        \resizebox{0.7\linewidth}{!}{
        \begin{tabular}{ l  c c}
    \toprule[1pt]
    Method & ins.\ mIoU & cat.\ mIoU \\
    \hline
    PointNet++~\cite{qi2017pointnet++}  & 85.1 & 81.9\\
    DGCNN~\cite{wang2019dgcnn}  & 85.2 & 82.3 \\%
    KPConv~\cite{thomas2019kpconv}& 86.4 & 85.1 \\
    PointMLP~\cite{ma2022pointmlp} & 86.1 & 84.6 \\ 
    Stratifiedformer~\cite{lai2022stratified} & 86.6 & 85.1 \\ 
    PTv1~\cite{zhao2021point} & 86.6 & 83.7 \\ 
    PointNeXt ~\cite{qian2022pointnext} & 87.0 & 85.2 \\
    Point-PN ~\cite{zhang2023starting} & 86.6 & - \\
    APES ~\cite{wu2023attention} & 85.8 & 83.7 \\
    ViPFormer~\cite{sun2023vipformer} & - & 84.7 \\
    Differ-GCN~\cite{gcn} & 86.5 & 84.5 \\
    ADS ~\cite{hong2023attention} & 86.9 & - \\
    SPoTr ~\cite{park2023self} & 87.2 & 85.4 \\
    \rowcolor{cyan! 20}AVS-Net (ours)  & \textbf{87.3} & \textbf{85.7} \\
    \bottomrule[1pt]
\end{tabular}

        }
        \label{tab:shapenetpartresult}
        \end{center}
    \end{minipage}
\end{table}

\begin{figure}
    \centering
    \includegraphics[width=0.49\textwidth]{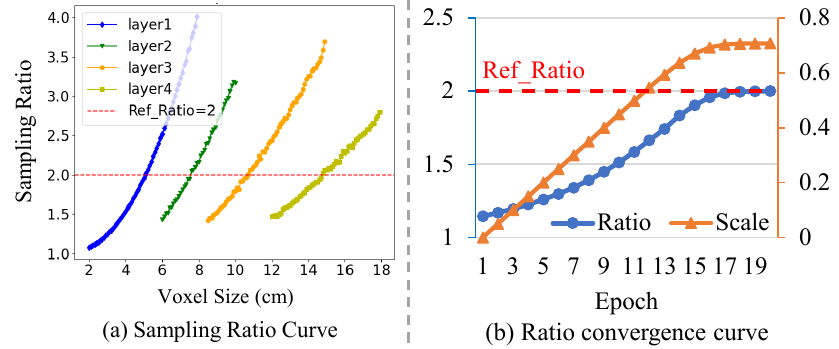}
        \caption{(a) The curve of the down-sampling ratio of the four-layer network as the voxel size changes. (b) Ratio convergence curve. The orange line represents the $Scale$ and the blue line represents the actual sampling ratio.}
      \vspace{-0.3cm}
    \label{fig:ratio_curve}
\end{figure}

\subsection{VAM Convergence Experiment}
We plot the curves of the down-sampling ratio with voxel size variation on the ShapeNetPart dataset. Different colors represent the four down-sampling layers of the network. As the voxel size increases, the average sampled points decrease, increasing the sampling rate. Fig.~\ref{fig:ratio_curve}(a) verifies that Eq.~\ref{eq:vsize_and_ratio2} is approximately valid.

On this basis, we adaptively adjust the voxel size using the PI control algorithm to match the given downsampling ratio, $Ref\_Ratio$. The scaling factor, $Scale$, is adjusted based on the actual sampling ratio. Fig.~\ref{fig:ratio_curve}(b) provides a specific example of how the scale changes. With a given downsampling ratio of $Ref\_Ratio=2$ and $\epsilon=0.1\%$, the initial voxel size, $V_0$, is set to $2.5cm$, and the initial $Scale$ is 0. Corresponding to this, the downsampling ratio is $1.14$. After $20$ epochs, the $Scale$ is adjusted to $0.708$, and according to Alg.~\ref{alg3}-\text{\romannumeral 4}, the sampled voxel size becomes $5.07cm$. At this point, the downsampling ratio is $1.9999$, meeting the termination condition in Alg.~\ref{alg3}. This experiment verifies that the PI control algorithm can make the downsampling ratio converge.

\subsection{Ablation Study}

\textbf{The Effectiveness of Each Component.}
We conducte ablation studies on the three modules of AVS-Net for 3D object detection and semantic segmentation tasks. The results for 3D detection are presented in Table~\ref{tab:waymo_ablation}. The baseline corresponds to the performance of CenterPoint-Pillar~\cite{yin2021center}. In Table~\ref{tab:waymo_ablation}(2), we only use the multi-layer Intra-VFE as the feature extraction module before BEV backbone. In Table~\ref{tab:waymo_ablation}(3) and (4), we introduce inter-VFE and VAM respectively, leading to a significant improvement in accuracy. Similar validations were conducted for the three modules in the semantic segmentation task. In Table~\ref{tab:VAM}(1), we utilize four layers of Intra-VFE for feature extraction. 
In Table~\ref{tab:VAM}(2), the Inter-VFE module is added to enable feature interaction between each voxel sampling point and its surrounding neighboring points. In Table~\ref{tab:VAM}(3), we use VAM to adaptively adjust the voxel size for each layer, resulting in a gain of $1.1\%$ in mIoU. The two sets of experiments in Table~\ref{tab:waymo_ablation} and Table~\ref{tab:VAM} validate the effectiveness of each proposed component.

\begin{table}
\begin{center}
\caption{{Ablation on each component. L1 and L2 Means LEVEL\_1 and LEVEL\_2 mAP on the Waymo validation set.}}
\resizebox{0.9\linewidth}{!}{
\begin{tabular}{cccccccc}
\hline
\multirow{2}{*}{ID} & \multirow{2}{*}{Method} & \multicolumn{2}{c}{Vehicle} & \multicolumn{2}{c}{Pedestrian} & \multicolumn{2}{c}{Cyclist} \\
&   & L1            & L2            & L1              & L2              & L1             & L2       \\ \hline
(1)                  & Baseline             & 73.4  & 65.1  & 75.4  & 67.6  & 67.8  & 65.3  \\
(2)                  & Intra-VFE            & 75.7  & 67.3  & 78.2  & 70.1  & 72.2  & 69.4   \\
(3)                  & (2) + Inter-VFE      & 77.1  & 68.8  & 81.4  & 73.6  & 75.4  & 72.7 \\
(4)                  & (3) + VAM            & \textbf{79.9}  & \textbf{71.6}  & \textbf{84.0}  & \textbf{76.4}  & \textbf{77.8}  & \textbf{75.0}   \\
                     \hline
\end{tabular}
}
\label{tab:waymo_ablation}
\end{center}
\vspace{-0.3cm}
\end{table}

\begin{table}[t]
\begin{center}
\caption{{Ablation on each component, evaluation on the ScanNet validation set.}}
\resizebox{0.8\linewidth}{!}{
\begin{tabular}{c c c c}
\hline
ID  & Method          & Latency (ms)    & mIoU (\%)  \\ \hline
(1) & Intra-VFE       & 71.8            & 72.4              \\
(2) & (1)+Inter-VFE   & 131.8           & 74.9              \\
(3) & (2)+VAM         & 120.4           & \textbf{76.0}           \\ \hline
\end{tabular}}
\label{tab:VAM}
\vspace{-0.5cm}
\end{center}
\end{table}

\begin{table}[t]
\begin{center}
\caption{Ablation on sampling method, evaluation on the ScanNet validation set.}
\resizebox{0.8\linewidth}{!}{
\begin{tabular}{c c c c}
\hline
ID  & Method          & Latency (ms)    & mIoU (\%)  \\ \hline
(1) & MinkowskiNet-34 & 87.8            & 72.2               \\
(2) & (1)+Intra-VFE   & 75.5            & 73.5               \\
(3) & (2)+VAM         & 74.9            & \textbf{75.7}      \\ \hline
\end{tabular}}
\label{tab:sampling}
\vspace{-0.5cm}
\end{center}
\end{table}

\textbf{Ablation on sampling method.}
To validate the generality of our sampling method, we also conduct experiments on the classic voxel-based network MinkowskiNet~\cite{choy20194d}. MinkowskiNet~\cite{choy20194d} voxelizes the point cloud and downsamples the voxels using ``stride $=2$" convolutional layers. MinkowskiNet does not support downsampling with the arbitrary ratio of voxel size. We replace these convolutional layers with our intra-VFE, and replace the upsampling deconvolution layers with trilinear interpolation, keeping the remaining parts unchanged. The experimental results are shown in Table~\ref{tab:sampling}. Due to the preservation of precise geometric information in the point cloud during downsampling by Intra-VFE, the performance is $1.3\%$ higher than the baseline with the quantization loss introduced by voxelization.
In Table~\ref{tab:sampling}(1) and (2), the initial voxel size of the first layer is set to $2$cm, and four downsampling layers increase the voxel size by a factor of $2$ at each layer. In Table~\ref{tab:sampling}(3), we use a downsampling rate of $4$ to adaptively adjust the voxel size for each layer, resulting in gains of $2.2\%$ of mIoU. Table~\ref{tab:sampling} proves that our downsampling algorithm can be used with other voxel-based networks and achieve significant performance improvements.

\textbf{Ablation on Ref\_Ratio in VAM.}
For downsampling methods, the downsampling ratio is a crucial parameter. The downsampling ratio based on FPS has undergone thorough ablation studies in point-based methods~\cite{qi2017pointnet++,zhao2021point,qian2022pointnext,zhang2023starting}. Therefore, the used downsampling ratio in point-based methods is of great reference value for our VAM.
Table~\ref{tab:ratio_and_size} illustrates the correspondence between $Ref\_Ratio$ and $voxel\_size$ on the ShapeNetPart dataset. Different $Ref\_Ratio$s are used for the four downsampling layers, resulting in various $voxel\_size$ determined by the VAM. Specific numerical values indicate that the ratio between $voxel\_size$ in subsequent layers is arbitrary. The first experiment is a reference experiment, and the voxel size is doubled in each layer.
On the ShapeNetPart dataset, the best results are obtained with a $Ref\_Ratio$ of 2. The optimal $Ref\_Ratio$ aligns with the downsampling ratio commonly used in the previous point-based methods~\cite{qi2017pointnet++,zhao2021point,qian2022pointnext,zhang2023starting}.

\begin{table}[t]
\centering
\caption{Voxel size \emph{w.r.t.} different Ref\_Ratio settings for $4$-layer network on ShapeNetPart.}
\resizebox{0.9\linewidth}{!}{
\begin{tabular}{c   c  c  c }
\hline
Ref\_Ratio & Voxel size (cm)  & ins. mIoU & cat. mIoU  \\  \hline
- & {[}5.00, 10.00,  20.00,   40.00{]}    & 86.8 & 84.8\\
1.5 & {[}3.89, 5.25,  6.57,   8.02{]}    & 87.2 & 85.5\\
2  &   {[}5.07, 7.54,  10.58,  14.58{]}   & \textbf{87.3} & \textbf{85.7} \\
3  &   {[}6.70, 11.62, 19.45,  35.14{]}   & 87.0 & 84.9\\
4  &   {[}7.89, 15.40, 31.22,  65.58{]}   & 86.9 & 84.7 \\
\hline
\end{tabular}
}
\label{tab:ratio_and_size}
\end{table}

\begin{table}[t!]
    \centering
        \caption{Latency (ms) of sampling method and neighbor search algorithm.}
    \resizebox{0.95\linewidth}{!}{
    \begin{tabular}{c c c c c c c}
        \hline
        \multirow{2}{*}{Method} & \multicolumn{6}{c}{Number of points} \\
        \cmidrule{2-7}
        & {10k} & {20k} & {40k} & 80k & 160k & 320k \\
        \hline
        {FPS}  & 7.59 & 23.71 & 82.49 & 315.11 & 1217.06 & 4805.92 \\
        {Intra-voxel Query} & 0.60 & 0.62 & 0.64 & 0.70 & 0.71 & 1.43 \\
        \hline
        {KNN}  & 2.16 & 8.24 & 32.63 & 130.06 & 519.03 & 2082.33 \\
        {Inter-voxel Query} & 1.41 & 1.42 & 1.69 & 2.76 & 6.08 & 12.06 \\
        \hline
    \end{tabular}}
    \label{tab:sample_time}
    \vspace{-5mm}
\end{table}

\textbf{Comparison on Latency of Different Sampling and Neighbor Search Methods.}
We compare the latency of different sampling methods and neighbor search methods. In Table~\ref{tab:sample_time}, both FPS and Intra-voxel Query are set with a downsampling ratio of $4$. KNN and Inter-voxel Query perform neighbor point searches on the downsampled point clouds. 
For the latency test of each algorithm, we averaged $100$ runs.
The time consumption of FPS and KNN increases quadratically with the number of points. In contrast, our proposed Intra-voxel and Inter-voxel Query algorithm show less noticeable changes when the point count is low due to the fixed time overhead of the $scatter$ operator. However, when the point count reaches $80k$, the time exhibits an apparent linear increase. For large-scale point clouds with number $\geq 320k$, our sampling method is $1000\times$ faster than FPS, and our neighbor search algorithm is $100\times$ faster than KNN.
\section{Conclusion}
\label{sec:Conclusion}

In this paper, to address the challenge of efficient sampling in large-scale point cloud scenes for autonomous driving, we propose an advanced sampler called Voxel Adaption Module (VAM). VAM can automatically adjust voxel size based on the downsampling ratio to achieve a favorable balance between efficiency and accuracy. We also construct the AVS-Net, which preserves geometric cues during sampling with efficient implementation of neighborhood search and feature aggregation. The proposed network can support our adaptive voxel size design well. Extensive experiments on multiple point cloud datasets have been conducted to demonstrate our method's superior accuracy and promising efficiency.






\bibliographystyle{cas-model2-names}

\bibliography{cas-refs}

\begin{thebibliography}{59}
\expandafter\ifx\csname natexlab\endcsname\relax\def\natexlab#1{#1}\fi
\providecommand{\url}[1]{\texttt{#1}}
\providecommand{\href}[2]{#2}
\providecommand{\path}[1]{#1}
\providecommand{\DOIprefix}{doi:}
\providecommand{\ArXivprefix}{arXiv:}
\providecommand{\URLprefix}{URL: }
\providecommand{\Pubmedprefix}{pmid:}
\providecommand{\doi}[1]{\href{http://dx.doi.org/#1}{\path{#1}}}
\providecommand{\Pubmed}[1]{\href{pmid:#1}{\path{#1}}}
\providecommand{\bibinfo}[2]{#2}
\ifx\xfnm\relax \def\xfnm[#1]{\unskip,\space#1}\fi
\bibitem[{Bai et~al.(2024)Bai, Li, Yang, Li, Xiao and Li}]{gcn}
\bibinfo{author}{Bai, Y.}, \bibinfo{author}{Li, G.}, \bibinfo{author}{Yang, C.}, \bibinfo{author}{Li, Y.}, \bibinfo{author}{Xiao, Q.}, \bibinfo{author}{Li, Z.}, \bibinfo{year}{2024}.
\newblock \bibinfo{title}{Differential graph convolution network for point cloud understanding}.
\newblock \bibinfo{journal}{Neurocomputing} , \bibinfo{pages}{127940}.
\bibitem[{Chen et~al.(2017)Chen, Ma, Wan, Li and Xia}]{chen2017multi}
\bibinfo{author}{Chen, X.}, \bibinfo{author}{Ma, H.}, \bibinfo{author}{Wan, J.}, \bibinfo{author}{Li, B.}, \bibinfo{author}{Xia, T.}, \bibinfo{year}{2017}.
\newblock \bibinfo{title}{Multi-view 3d object detection network for autonomous driving}, in: \bibinfo{booktitle}{IEEE Conf. Comput. Vis. Pattern Recog.}, pp. \bibinfo{pages}{1907--1915}.
\bibitem[{Chen et~al.(2023a)Chen, Liu, Zhang, Qi and Jia}]{chen2023largekernel3d}
\bibinfo{author}{Chen, Y.}, \bibinfo{author}{Liu, J.}, \bibinfo{author}{Zhang, X.}, \bibinfo{author}{Qi, X.}, \bibinfo{author}{Jia, J.}, \bibinfo{year}{2023}a.
\newblock \bibinfo{title}{Largekernel3d: Scaling up kernels in 3d sparse cnns}, in: \bibinfo{booktitle}{IEEE Conf. Comput. Vis. Pattern Recog.}, pp. \bibinfo{pages}{13488--13498}.
\bibitem[{Chen et~al.(2023b)Chen, Liu, Zhang, Qi and Jia}]{chen2023voxelnext}
\bibinfo{author}{Chen, Y.}, \bibinfo{author}{Liu, J.}, \bibinfo{author}{Zhang, X.}, \bibinfo{author}{Qi, X.}, \bibinfo{author}{Jia, J.}, \bibinfo{year}{2023}b.
\newblock \bibinfo{title}{Voxelnext: Fully sparse voxelnet for 3d object detection and tracking}, in: \bibinfo{booktitle}{IEEE Conf. Comput. Vis. Pattern Recog.}, pp. \bibinfo{pages}{21674--21683}.
\bibitem[{Choy et~al.(2019)Choy, Gwak and Savarese}]{choy20194d}
\bibinfo{author}{Choy, C.}, \bibinfo{author}{Gwak, J.}, \bibinfo{author}{Savarese, S.}, \bibinfo{year}{2019}.
\newblock \bibinfo{title}{4d spatio-temporal convnets: Minkowski convolutional neural networks}, in: \bibinfo{booktitle}{IEEE Conf. Comput. Vis. Pattern Recog.}, pp. \bibinfo{pages}{3075--3084}.
\bibitem[{Dai et~al.(2017)Dai, Chang, Savva, Halber, Funkhouser and Nie{\ss}ner}]{dai2017scannet}
\bibinfo{author}{Dai, A.}, \bibinfo{author}{Chang, A.X.}, \bibinfo{author}{Savva, M.}, \bibinfo{author}{Halber, M.}, \bibinfo{author}{Funkhouser, T.}, \bibinfo{author}{Nie{\ss}ner, M.}, \bibinfo{year}{2017}.
\newblock \bibinfo{title}{{ScanNet}: Richly-annotated {3D} reconstructions of indoor scenes}, in: \bibinfo{booktitle}{IEEE Conf. Comput. Vis. Pattern Recog.}
\bibitem[{Deng et~al.(2021)Deng, Shi, Li, Zhou, Zhang and Li}]{deng2021voxel}
\bibinfo{author}{Deng, J.}, \bibinfo{author}{Shi, S.}, \bibinfo{author}{Li, P.}, \bibinfo{author}{Zhou, W.}, \bibinfo{author}{Zhang, Y.}, \bibinfo{author}{Li, H.}, \bibinfo{year}{2021}.
\newblock \bibinfo{title}{Voxel r-cnn: Towards high performance voxel-based 3d object detection}, in: \bibinfo{booktitle}{Proc. AAAI Conf. Artif. Intell.}, pp. \bibinfo{pages}{1201--1209}.
\bibitem[{Dovrat et~al.(2019)Dovrat, Lang and Avidan}]{dovrat2019learning}
\bibinfo{author}{Dovrat, O.}, \bibinfo{author}{Lang, I.}, \bibinfo{author}{Avidan, S.}, \bibinfo{year}{2019}.
\newblock \bibinfo{title}{Learning to sample}, in: \bibinfo{booktitle}{IEEE Conf. Comput. Vis. Pattern Recog.}, pp. \bibinfo{pages}{2760--2769}.
\bibitem[{Fan et~al.(2022)Fan, Pang, Zhang, Wang, Zhao, Wang, Wang and Zhang}]{fan2021embracing}
\bibinfo{author}{Fan, L.}, \bibinfo{author}{Pang, Z.}, \bibinfo{author}{Zhang, T.}, \bibinfo{author}{Wang, Y.X.}, \bibinfo{author}{Zhao, H.}, \bibinfo{author}{Wang, F.}, \bibinfo{author}{Wang, N.}, \bibinfo{author}{Zhang, Z.}, \bibinfo{year}{2022}.
\newblock \bibinfo{title}{Embracing single stride 3d object detector with sparse transformer}, in: \bibinfo{booktitle}{IEEE Conf. Comput. Vis. Pattern Recog.}
\bibitem[{Graham et~al.(2018)Graham, Engelcke and Van Der~Maaten}]{graham20183d}
\bibinfo{author}{Graham, B.}, \bibinfo{author}{Engelcke, M.}, \bibinfo{author}{Van Der~Maaten, L.}, \bibinfo{year}{2018}.
\newblock \bibinfo{title}{3d semantic segmentation with submanifold sparse convolutional networks}, in: \bibinfo{booktitle}{IEEE Conf. Comput. Vis. Pattern Recog.}, pp. \bibinfo{pages}{9224--9232}.
\bibitem[{He et~al.(2022a)He, Li, Li and Zhang}]{he2022voxel}
\bibinfo{author}{He, C.}, \bibinfo{author}{Li, R.}, \bibinfo{author}{Li, S.}, \bibinfo{author}{Zhang, L.}, \bibinfo{year}{2022}a.
\newblock \bibinfo{title}{Voxel set transformer: A set-to-set approach to 3d object detection from point clouds}, in: \bibinfo{booktitle}{IEEE Conf. Comput. Vis. Pattern Recog.}, pp. \bibinfo{pages}{8417--8427}.
\bibitem[{He et~al.(2022b)He, Li, Li and Zhang}]{voxelset}
\bibinfo{author}{He, C.}, \bibinfo{author}{Li, R.}, \bibinfo{author}{Li, S.}, \bibinfo{author}{Zhang, L.}, \bibinfo{year}{2022}b.
\newblock \bibinfo{title}{Voxel set transformer: A set-to-set approach to 3d object detection from point clouds}, in: \bibinfo{booktitle}{IEEE Conf. Comput. Vis. Pattern Recog.}
\bibitem[{Hong et~al.(2023)Hong, Chou and Liu}]{hong2023attention}
\bibinfo{author}{Hong, C.Y.}, \bibinfo{author}{Chou, Y.Y.}, \bibinfo{author}{Liu, T.L.}, \bibinfo{year}{2023}.
\newblock \bibinfo{title}{Attention discriminant sampling for point clouds}, in: \bibinfo{booktitle}{Int. Conf. Comput. Vis.}, pp. \bibinfo{pages}{14429--14440}.
\bibitem[{Hu et~al.(2022)Hu, Kuai and Waslander}]{hu2022point}
\bibinfo{author}{Hu, J.S.}, \bibinfo{author}{Kuai, T.}, \bibinfo{author}{Waslander, S.L.}, \bibinfo{year}{2022}.
\newblock \bibinfo{title}{Point density-aware voxels for lidar 3d object detection}, in: \bibinfo{booktitle}{IEEE Conf. Comput. Vis. Pattern Recog.}, pp. \bibinfo{pages}{8469--8478}.
\bibitem[{Hu et~al.(2020)Hu, Yang, Xie, Rosa, Guo, Wang, Trigoni and Markham}]{hu2020randla}
\bibinfo{author}{Hu, Q.}, \bibinfo{author}{Yang, B.}, \bibinfo{author}{Xie, L.}, \bibinfo{author}{Rosa, S.}, \bibinfo{author}{Guo, Y.}, \bibinfo{author}{Wang, Z.}, \bibinfo{author}{Trigoni, N.}, \bibinfo{author}{Markham, A.}, \bibinfo{year}{2020}.
\newblock \bibinfo{title}{Randla-net: Efficient semantic segmentation of large-scale point clouds}, in: \bibinfo{booktitle}{IEEE Conf. Comput. Vis. Pattern Recog.}
\bibitem[{Huang et~al.(2024a)Huang, Wang, Yang, Yang, Lu, Chen, Yu and Zhang}]{iefm}
\bibinfo{author}{Huang, K.}, \bibinfo{author}{Wang, J.}, \bibinfo{author}{Yang, J.}, \bibinfo{author}{Yang, Y.}, \bibinfo{author}{Lu, G.}, \bibinfo{author}{Chen, Y.}, \bibinfo{author}{Yu, H.}, \bibinfo{author}{Zhang, Q.}, \bibinfo{year}{2024}a.
\newblock \bibinfo{title}{Iefm and ids: Enhancing 3d environment perception via information encoding in indoor point cloud semantic segmentation}.
\newblock \bibinfo{journal}{Neurocomputing} \bibinfo{volume}{563}, \bibinfo{pages}{126944}.
\bibitem[{Huang et~al.(2024b)Huang, Zhou, Yan and Zheng}]{kpdet}
\bibinfo{author}{Huang, Y.}, \bibinfo{author}{Zhou, S.}, \bibinfo{author}{Yan, X.}, \bibinfo{author}{Zheng, N.}, \bibinfo{year}{2024}b.
\newblock \bibinfo{title}{Kpdet: Keypoint-based 3d object detection with parametric radius learning}.
\newblock \bibinfo{journal}{Neurocomputing} \bibinfo{volume}{572}, \bibinfo{pages}{127171}.
\bibitem[{Lai et~al.(2022)Lai, Liu, Jiang, Wang, Zhao, Liu, Qi and Jia}]{lai2022stratified}
\bibinfo{author}{Lai, X.}, \bibinfo{author}{Liu, J.}, \bibinfo{author}{Jiang, L.}, \bibinfo{author}{Wang, L.}, \bibinfo{author}{Zhao, H.}, \bibinfo{author}{Liu, S.}, \bibinfo{author}{Qi, X.}, \bibinfo{author}{Jia, J.}, \bibinfo{year}{2022}.
\newblock \bibinfo{title}{Stratified transformer for 3d point cloud segmentation}, in: \bibinfo{booktitle}{IEEE Conf. Comput. Vis. Pattern Recog.}
\bibitem[{Lang et~al.(2019)Lang, Vora, Caesar, Zhou, Yang and Beijbom}]{lang2019pointpillars}
\bibinfo{author}{Lang, A.H.}, \bibinfo{author}{Vora, S.}, \bibinfo{author}{Caesar, H.}, \bibinfo{author}{Zhou, L.}, \bibinfo{author}{Yang, J.}, \bibinfo{author}{Beijbom, O.}, \bibinfo{year}{2019}.
\newblock \bibinfo{title}{Pointpillars: Fast encoders for object detection from point clouds}, in: \bibinfo{booktitle}{IEEE Conf. Comput. Vis. Pattern Recog.}, pp. \bibinfo{pages}{12697--12705}.
\bibitem[{Liu et~al.(2019)Liu, Tang, Lin and Han}]{liu2019pvcnn}
\bibinfo{author}{Liu, Z.}, \bibinfo{author}{Tang, H.}, \bibinfo{author}{Lin, Y.}, \bibinfo{author}{Han, S.}, \bibinfo{year}{2019}.
\newblock \bibinfo{title}{Point-voxel cnn for efficient 3d deep learning}, in: \bibinfo{booktitle}{Adv. Neural Inform. Process. Syst.}
\bibitem[{Liu et~al.(2023)Liu, Yang, Tang, Yang and Han}]{liu2023flatformer}
\bibinfo{author}{Liu, Z.}, \bibinfo{author}{Yang, X.}, \bibinfo{author}{Tang, H.}, \bibinfo{author}{Yang, S.}, \bibinfo{author}{Han, S.}, \bibinfo{year}{2023}.
\newblock \bibinfo{title}{Flatformer: Flattened window attention for efficient point cloud transformer}, in: \bibinfo{booktitle}{IEEE Conf. Comput. Vis. Pattern Recog.}, pp. \bibinfo{pages}{1200--1211}.
\bibitem[{Lu et~al.(2023)Lu, Zhao, Premebida, Zhang, Zhao and Tian}]{lu2023improving}
\bibinfo{author}{Lu, W.}, \bibinfo{author}{Zhao, D.}, \bibinfo{author}{Premebida, C.}, \bibinfo{author}{Zhang, L.}, \bibinfo{author}{Zhao, W.}, \bibinfo{author}{Tian, D.}, \bibinfo{year}{2023}.
\newblock \bibinfo{title}{Improving 3d vulnerable road user detection with point augmentation}.
\newblock \bibinfo{journal}{IEEE Trans. Intell. Veh.} .
\bibitem[{Ma et~al.(2022)Ma, Qin, You, Ran and Fu}]{ma2022pointmlp}
\bibinfo{author}{Ma, X.}, \bibinfo{author}{Qin, C.}, \bibinfo{author}{You, H.}, \bibinfo{author}{Ran, H.}, \bibinfo{author}{Fu, Y.}, \bibinfo{year}{2022}.
\newblock \bibinfo{title}{Rethinking network design and local geometry in point cloud: A simple residual {MLP} framework}, in: \bibinfo{booktitle}{Int. Conf. Learn. Represent.}
\bibitem[{Ouyang et~al.(2023)Ouyang, Liu and Chen}]{ouyang2023hierarchical}
\bibinfo{author}{Ouyang, J.}, \bibinfo{author}{Liu, X.}, \bibinfo{author}{Chen, H.}, \bibinfo{year}{2023}.
\newblock \bibinfo{title}{Hierarchical adaptive voxel-guided sampling for real-time applications in large-scale point clouds}.
\newblock \bibinfo{journal}{arXiv preprint arXiv:2305.14306} .
\bibitem[{Park et~al.(2022)Park, Jeong, Cho and Park}]{park2022fast}
\bibinfo{author}{Park, C.}, \bibinfo{author}{Jeong, Y.}, \bibinfo{author}{Cho, M.}, \bibinfo{author}{Park, J.}, \bibinfo{year}{2022}.
\newblock \bibinfo{title}{Fast point transformer}, in: \bibinfo{booktitle}{IEEE Conf. Comput. Vis. Pattern Recog.}, pp. \bibinfo{pages}{16949--16958}.
\bibitem[{Park et~al.(2023)Park, Lee, Kim, Xiong and Kim}]{park2023self}
\bibinfo{author}{Park, J.}, \bibinfo{author}{Lee, S.}, \bibinfo{author}{Kim, S.}, \bibinfo{author}{Xiong, Y.}, \bibinfo{author}{Kim, H.J.}, \bibinfo{year}{2023}.
\newblock \bibinfo{title}{Self-positioning point-based transformer for point cloud understanding}, in: \bibinfo{booktitle}{IEEE Conf. Comput. Vis. Pattern Recog.}, pp. \bibinfo{pages}{21814--21823}.
\bibitem[{Qi et~al.(2017)Qi, Yi, Su and Guibas}]{qi2017pointnet++}
\bibinfo{author}{Qi, C.R.}, \bibinfo{author}{Yi, L.}, \bibinfo{author}{Su, H.}, \bibinfo{author}{Guibas, L.J.}, \bibinfo{year}{2017}.
\newblock \bibinfo{title}{Pointnet++: Deep hierarchical feature learning on point sets in a metric space}.
\newblock \bibinfo{journal}{Adv. Neural Inform. Process. Syst.} \bibinfo{volume}{30}.
\bibitem[{Qian et~al.(2022)Qian, Li, Peng, Mai, Hammoud, Elhoseiny and Ghanem}]{qian2022pointnext}
\bibinfo{author}{Qian, G.}, \bibinfo{author}{Li, Y.}, \bibinfo{author}{Peng, H.}, \bibinfo{author}{Mai, J.}, \bibinfo{author}{Hammoud, H.}, \bibinfo{author}{Elhoseiny, M.}, \bibinfo{author}{Ghanem, B.}, \bibinfo{year}{2022}.
\newblock \bibinfo{title}{Pointnext: Revisiting pointnet++ with improved training and scaling strategies}.
\newblock \bibinfo{journal}{Adv. Neural Inform. Process. Syst.} \bibinfo{volume}{35}, \bibinfo{pages}{23192--23204}.
\bibitem[{Sheng et~al.(2021)Sheng, Cai, Liu, Deng, Huang, Hua and Zhao}]{sheng2021improving}
\bibinfo{author}{Sheng, H.}, \bibinfo{author}{Cai, S.}, \bibinfo{author}{Liu, Y.}, \bibinfo{author}{Deng, B.}, \bibinfo{author}{Huang, J.}, \bibinfo{author}{Hua, X.S.}, \bibinfo{author}{Zhao, M.J.}, \bibinfo{year}{2021}.
\newblock \bibinfo{title}{Improving 3d object detection with channel-wise transformer}, in: \bibinfo{booktitle}{Int. Conf. Comput. Vis.}, pp. \bibinfo{pages}{2743--2752}.
\bibitem[{Shi et~al.(2022a)Shi, Li and Ma}]{shi2022pillarnet}
\bibinfo{author}{Shi, G.}, \bibinfo{author}{Li, R.}, \bibinfo{author}{Ma, C.}, \bibinfo{year}{2022}a.
\newblock \bibinfo{title}{Pillarnet: Real-time and high-performance pillar-based 3d object detection}, in: \bibinfo{booktitle}{Eur. Conf. Comput. Vis.}
\bibitem[{Shi et~al.(2020a)Shi, Guo, Jiang, Wang, Shi, Wang and Li}]{shi2020pv}
\bibinfo{author}{Shi, S.}, \bibinfo{author}{Guo, C.}, \bibinfo{author}{Jiang, L.}, \bibinfo{author}{Wang, Z.}, \bibinfo{author}{Shi, J.}, \bibinfo{author}{Wang, X.}, \bibinfo{author}{Li, H.}, \bibinfo{year}{2020}a.
\newblock \bibinfo{title}{Pv-rcnn: Point-voxel feature set abstraction for 3d object detection}, in: \bibinfo{booktitle}{IEEE Conf. Comput. Vis. Pattern Recog.}, pp. \bibinfo{pages}{10529--10538}.
\bibitem[{Shi et~al.(2022b)Shi, Jiang, Deng, Wang, Guo, Shi, Wang and Li}]{shi2022pv}
\bibinfo{author}{Shi, S.}, \bibinfo{author}{Jiang, L.}, \bibinfo{author}{Deng, J.}, \bibinfo{author}{Wang, Z.}, \bibinfo{author}{Guo, C.}, \bibinfo{author}{Shi, J.}, \bibinfo{author}{Wang, X.}, \bibinfo{author}{Li, H.}, \bibinfo{year}{2022}b.
\newblock \bibinfo{title}{Pv-rcnn++: Point-voxel feature set abstraction with local vector representation for 3d object detection}.
\newblock \bibinfo{journal}{Int. J. Comput. Vis.} , \bibinfo{pages}{1--21}.
\bibitem[{Shi et~al.(2019)Shi, Wang and Li}]{shi2019pointrcnn}
\bibinfo{author}{Shi, S.}, \bibinfo{author}{Wang, X.}, \bibinfo{author}{Li, H.}, \bibinfo{year}{2019}.
\newblock \bibinfo{title}{Pointrcnn: 3d object proposal generation and detection from point cloud}, in: \bibinfo{booktitle}{IEEE Conf. Comput. Vis. Pattern Recog.}, pp. \bibinfo{pages}{770--779}.
\bibitem[{Shi et~al.(2020b)Shi, Wang, Shi, Wang and Li}]{shi2020points}
\bibinfo{author}{Shi, S.}, \bibinfo{author}{Wang, Z.}, \bibinfo{author}{Shi, J.}, \bibinfo{author}{Wang, X.}, \bibinfo{author}{Li, H.}, \bibinfo{year}{2020}b.
\newblock \bibinfo{title}{From points to parts: 3d object detection from point cloud with part-aware and part-aggregation network}.
\newblock \bibinfo{journal}{IEEE Trans. Pattern Anal. Mach. Intell.} \bibinfo{volume}{43}, \bibinfo{pages}{2647--2664}.
\bibitem[{Su et~al.(2015)Su, Maji, Kalogerakis and Learned{-}Miller}]{su15mvcnn}
\bibinfo{author}{Su, H.}, \bibinfo{author}{Maji, S.}, \bibinfo{author}{Kalogerakis, E.}, \bibinfo{author}{Learned{-}Miller, E.G.}, \bibinfo{year}{2015}.
\newblock \bibinfo{title}{Multi-view convolutional neural networks for 3d shape recognition}, in: \bibinfo{booktitle}{Int. Conf. Comput. Vis.}
\bibitem[{Sun et~al.(2023)Sun, Wang, Cai, Bai and Li}]{sun2023vipformer}
\bibinfo{author}{Sun, H.}, \bibinfo{author}{Wang, Y.}, \bibinfo{author}{Cai, X.}, \bibinfo{author}{Bai, X.}, \bibinfo{author}{Li, D.}, \bibinfo{year}{2023}.
\newblock \bibinfo{title}{Vipformer: Efficient vision-and-pointcloud transformer for unsupervised pointcloud understanding}, in: \bibinfo{booktitle}{2023 IEEE International Conference on Robotics and Automation (ICRA)}, \bibinfo{organization}{IEEE}. pp. \bibinfo{pages}{7234--7242}.
\bibitem[{Sun et~al.(2020)Sun, Kretzschmar, Dotiwalla, Chouard, Patnaik, Tsui, Guo, Zhou, Chai, Caine et~al.}]{sun2020scalability}
\bibinfo{author}{Sun, P.}, \bibinfo{author}{Kretzschmar, H.}, \bibinfo{author}{Dotiwalla, X.}, \bibinfo{author}{Chouard, A.}, \bibinfo{author}{Patnaik, V.}, \bibinfo{author}{Tsui, P.}, \bibinfo{author}{Guo, J.}, \bibinfo{author}{Zhou, Y.}, \bibinfo{author}{Chai, Y.}, \bibinfo{author}{Caine, B.}, et~al., \bibinfo{year}{2020}.
\newblock \bibinfo{title}{Scalability in perception for autonomous driving: Waymo open dataset}, in: \bibinfo{booktitle}{IEEE Conf. Comput. Vis. Pattern Recog.}, pp. \bibinfo{pages}{2446--2454}.
\bibitem[{Tang et~al.(2020)Tang, Liu, Zhao, Lin, Lin, Wang and Han}]{tang2020searching}
\bibinfo{author}{Tang, H.}, \bibinfo{author}{Liu, Z.}, \bibinfo{author}{Zhao, S.}, \bibinfo{author}{Lin, Y.}, \bibinfo{author}{Lin, J.}, \bibinfo{author}{Wang, H.}, \bibinfo{author}{Han, S.}, \bibinfo{year}{2020}.
\newblock \bibinfo{title}{Searching efficient 3d architectures with sparse point-voxel convolution}, in: \bibinfo{booktitle}{Eur. Conf. Comput. Vis.}, \bibinfo{organization}{Springer}. pp. \bibinfo{pages}{685--702}.
\bibitem[{Tang et~al.(2023)Tang, He, Wang, Mao and Wang}]{tang2023multi}
\bibinfo{author}{Tang, Y.}, \bibinfo{author}{He, H.}, \bibinfo{author}{Wang, Y.}, \bibinfo{author}{Mao, Z.}, \bibinfo{author}{Wang, H.}, \bibinfo{year}{2023}.
\newblock \bibinfo{title}{Multi-modality 3d object detection in autonomous driving: A review}.
\newblock \bibinfo{journal}{Neurocomputing} , \bibinfo{pages}{126587}.
\bibitem[{Thomas et~al.(2019)Thomas, Qi, Deschaud, Marcotegui, Goulette and Guibas}]{thomas2019kpconv}
\bibinfo{author}{Thomas, H.}, \bibinfo{author}{Qi, C.R.}, \bibinfo{author}{Deschaud, J.E.}, \bibinfo{author}{Marcotegui, B.}, \bibinfo{author}{Goulette, F.}, \bibinfo{author}{Guibas, L.J.}, \bibinfo{year}{2019}.
\newblock \bibinfo{title}{Kpconv: Flexible and deformable convolution for point clouds}, in: \bibinfo{booktitle}{Int. Conf. Comput. Vis.}
\bibitem[{Vaswani et~al.(2017)Vaswani, Shazeer, Parmar, Uszkoreit, Jones, Gomez, Kaiser and Polosukhin}]{vaswani2017attention}
\bibinfo{author}{Vaswani, A.}, \bibinfo{author}{Shazeer, N.}, \bibinfo{author}{Parmar, N.}, \bibinfo{author}{Uszkoreit, J.}, \bibinfo{author}{Jones, L.}, \bibinfo{author}{Gomez, A.N.}, \bibinfo{author}{Kaiser, {\L}.}, \bibinfo{author}{Polosukhin, I.}, \bibinfo{year}{2017}.
\newblock \bibinfo{title}{Attention is all you need}.
\newblock \bibinfo{journal}{Adv. Neural Inform. Process. Syst.} \bibinfo{volume}{30}.
\bibitem[{Wang et~al.(2023)Wang, Shi, Shi, Lei, Wang, He, Schiele and Wang}]{DSVT}
\bibinfo{author}{Wang, H.}, \bibinfo{author}{Shi, C.}, \bibinfo{author}{Shi, S.}, \bibinfo{author}{Lei, M.}, \bibinfo{author}{Wang, S.}, \bibinfo{author}{He, D.}, \bibinfo{author}{Schiele, B.}, \bibinfo{author}{Wang, L.}, \bibinfo{year}{2023}.
\newblock \bibinfo{title}{Dsvt: Dynamic sparse voxel transformer with rotated sets}, in: \bibinfo{booktitle}{IEEE Conf. Comput. Vis. Pattern Recog.}
\bibitem[{Wang(2023)}]{Wang2023OctFormer}
\bibinfo{author}{Wang, P.S.}, \bibinfo{year}{2023}.
\newblock \bibinfo{title}{Octformer: Octree-based transformers for 3d point clouds}.
\newblock \bibinfo{journal}{ACM Trans. Graph.} \bibinfo{volume}{42}, \bibinfo{pages}{1--11}.
\bibitem[{Wang et~al.(2019)Wang, Sun, Liu, Sarma, Bronstein and Solomon}]{wang2019dgcnn}
\bibinfo{author}{Wang, Y.}, \bibinfo{author}{Sun, Y.}, \bibinfo{author}{Liu, Z.}, \bibinfo{author}{Sarma, S.E.}, \bibinfo{author}{Bronstein, M.M.}, \bibinfo{author}{Solomon, J.M.}, \bibinfo{year}{2019}.
\newblock \bibinfo{title}{Dynamic graph cnn for learning on point clouds}.
\newblock \bibinfo{journal}{ACM Trans. Graph.} .
\bibitem[{Wu et~al.(2023)Wu, Zheng, Pfrommer and Beyerer}]{wu2023attention}
\bibinfo{author}{Wu, C.}, \bibinfo{author}{Zheng, J.}, \bibinfo{author}{Pfrommer, J.}, \bibinfo{author}{Beyerer, J.}, \bibinfo{year}{2023}.
\newblock \bibinfo{title}{Attention-based point cloud edge sampling}, in: \bibinfo{booktitle}{IEEE Conf. Comput. Vis. Pattern Recog.}, pp. \bibinfo{pages}{5333--5343}.
\bibitem[{Wu et~al.(2019)Wu, Qi and Fuxin}]{wu2019pointconv}
\bibinfo{author}{Wu, W.}, \bibinfo{author}{Qi, Z.}, \bibinfo{author}{Fuxin, L.}, \bibinfo{year}{2019}.
\newblock \bibinfo{title}{Pointconv: Deep convolutional networks on 3d point clouds}, in: \bibinfo{booktitle}{IEEE Conf. Comput. Vis. Pattern Recog.}
\bibitem[{Wu et~al.(2022)Wu, Lao, Jiang, Liu and Zhao}]{wu2022point}
\bibinfo{author}{Wu, X.}, \bibinfo{author}{Lao, Y.}, \bibinfo{author}{Jiang, L.}, \bibinfo{author}{Liu, X.}, \bibinfo{author}{Zhao, H.}, \bibinfo{year}{2022}.
\newblock \bibinfo{title}{Point transformer v2: Grouped vector attention and partition-based pooling}.
\newblock \bibinfo{journal}{Adv. Neural Inform. Process. Syst.} \bibinfo{volume}{35}, \bibinfo{pages}{33330--33342}.
\bibitem[{Xu et~al.(2022)Xu, Zhong and Neumann}]{xu2022behind}
\bibinfo{author}{Xu, Q.}, \bibinfo{author}{Zhong, Y.}, \bibinfo{author}{Neumann, U.}, \bibinfo{year}{2022}.
\newblock \bibinfo{title}{Behind the curtain: Learning occluded shapes for 3d object detection}, in: \bibinfo{booktitle}{Proc. AAAI Conf. Artif. Intell.}, pp. \bibinfo{pages}{2893--2901}.
\bibitem[{Yan et~al.(2020)Yan, Zheng, Li, Wang and Cui}]{yan2020pointasnl}
\bibinfo{author}{Yan, X.}, \bibinfo{author}{Zheng, C.}, \bibinfo{author}{Li, Z.}, \bibinfo{author}{Wang, S.}, \bibinfo{author}{Cui, S.}, \bibinfo{year}{2020}.
\newblock \bibinfo{title}{Pointasnl: Robust point clouds processing using nonlocal neural networks with adaptive sampling}, in: \bibinfo{booktitle}{IEEE Conf. Comput. Vis. Pattern Recog.}
\bibitem[{Yan et~al.(2018)Yan, Mao and Li}]{yan2018second}
\bibinfo{author}{Yan, Y.}, \bibinfo{author}{Mao, Y.}, \bibinfo{author}{Li, B.}, \bibinfo{year}{2018}.
\newblock \bibinfo{title}{Second: Sparsely embedded convolutional detection}.
\newblock \bibinfo{journal}{Sensors} \bibinfo{volume}{18}, \bibinfo{pages}{3337}.
\bibitem[{Yang et~al.(2023)Yang, Song, Liu, Mao, Li, Li, Sun, Sun and Zheng}]{dbqssd}
\bibinfo{author}{Yang, J.}, \bibinfo{author}{Song, L.}, \bibinfo{author}{Liu, S.}, \bibinfo{author}{Mao, W.}, \bibinfo{author}{Li, Z.}, \bibinfo{author}{Li, X.}, \bibinfo{author}{Sun, H.}, \bibinfo{author}{Sun, J.}, \bibinfo{author}{Zheng, N.}, \bibinfo{year}{2023}.
\newblock \bibinfo{title}{Dbq-ssd: Dynamic ball query for efficient 3d object detection}, in: \bibinfo{booktitle}{Int. Conf. Learn. Represent.}
\bibitem[{Yi et~al.(2016)Yi, Kim, Ceylan, Shen, Yan, Su, Lu, Huang, Sheffer and Guibas}]{yi2016scalable}
\bibinfo{author}{Yi, L.}, \bibinfo{author}{Kim, V.G.}, \bibinfo{author}{Ceylan, D.}, \bibinfo{author}{Shen, I.C.}, \bibinfo{author}{Yan, M.}, \bibinfo{author}{Su, H.}, \bibinfo{author}{Lu, C.}, \bibinfo{author}{Huang, Q.}, \bibinfo{author}{Sheffer, A.}, \bibinfo{author}{Guibas, L.}, \bibinfo{year}{2016}.
\newblock \bibinfo{title}{A scalable active framework for region annotation in 3d shape collections}.
\newblock \bibinfo{journal}{ACM Trans. Graph.} .
\bibitem[{Yin et~al.(2021)Yin, Zhou and Krahenbuhl}]{yin2021center}
\bibinfo{author}{Yin, T.}, \bibinfo{author}{Zhou, X.}, \bibinfo{author}{Krahenbuhl, P.}, \bibinfo{year}{2021}.
\newblock \bibinfo{title}{Center-based 3d object detection and tracking}, in: \bibinfo{booktitle}{IEEE Conf. Comput. Vis. Pattern Recog.}, pp. \bibinfo{pages}{11784--11793}.
\bibitem[{Zhang et~al.(2022a)Zhang, Wan, Shen and Wu}]{zhang2022pvt}
\bibinfo{author}{Zhang, C.}, \bibinfo{author}{Wan, H.}, \bibinfo{author}{Shen, X.}, \bibinfo{author}{Wu, Z.}, \bibinfo{year}{2022}a.
\newblock \bibinfo{title}{Pvt: Point-voxel transformer for point cloud learning}.
\newblock \bibinfo{journal}{International Journal of Intelligent Systems} \bibinfo{volume}{37}, \bibinfo{pages}{11985--12008}.
\bibitem[{Zhang et~al.(2020)Zhang, Fang, Wah and Torr}]{zhang2020deep}
\bibinfo{author}{Zhang, F.}, \bibinfo{author}{Fang, J.}, \bibinfo{author}{Wah, B.}, \bibinfo{author}{Torr, P.}, \bibinfo{year}{2020}.
\newblock \bibinfo{title}{Deep fusionnet for point cloud semantic segmentation}, in: \bibinfo{booktitle}{Eur. Conf. Comput. Vis.}, \bibinfo{organization}{Springer}. pp. \bibinfo{pages}{644--663}.
\bibitem[{Zhang et~al.(2023)Zhang, Wang, Wang, Gao, Li and Shi}]{zhang2023starting}
\bibinfo{author}{Zhang, R.}, \bibinfo{author}{Wang, L.}, \bibinfo{author}{Wang, Y.}, \bibinfo{author}{Gao, P.}, \bibinfo{author}{Li, H.}, \bibinfo{author}{Shi, J.}, \bibinfo{year}{2023}.
\newblock \bibinfo{title}{Starting from non-parametric networks for 3d point cloud analysis}, in: \bibinfo{booktitle}{IEEE Conf. Comput. Vis. Pattern Recog.}, pp. \bibinfo{pages}{5344--5353}.
\bibitem[{Zhang et~al.(2022b)Zhang, Hu, Xu, Ma, Wan and Guo}]{zhang2022not}
\bibinfo{author}{Zhang, Y.}, \bibinfo{author}{Hu, Q.}, \bibinfo{author}{Xu, G.}, \bibinfo{author}{Ma, Y.}, \bibinfo{author}{Wan, J.}, \bibinfo{author}{Guo, Y.}, \bibinfo{year}{2022}b.
\newblock \bibinfo{title}{Not all points are equal: Learning highly efficient point-based detectors for 3d lidar point clouds}, in: \bibinfo{booktitle}{IEEE Conf. Comput. Vis. Pattern Recog.}, pp. \bibinfo{pages}{18953--18962}.
\bibitem[{Zhao et~al.(2021)Zhao, Jiang, Jia, Torr and Koltun}]{zhao2021point}
\bibinfo{author}{Zhao, H.}, \bibinfo{author}{Jiang, L.}, \bibinfo{author}{Jia, J.}, \bibinfo{author}{Torr, P.H.}, \bibinfo{author}{Koltun, V.}, \bibinfo{year}{2021}.
\newblock \bibinfo{title}{Point transformer}, in: \bibinfo{booktitle}{Int. Conf. Comput. Vis.}, pp. \bibinfo{pages}{16259--16268}.
\bibitem[{Zhou et~al.(2022)Zhou, Zhao, Wang, Wang and Foroosh}]{Zhou_centerformer}
\bibinfo{author}{Zhou, Z.}, \bibinfo{author}{Zhao, X.}, \bibinfo{author}{Wang, Y.}, \bibinfo{author}{Wang, P.}, \bibinfo{author}{Foroosh, H.}, \bibinfo{year}{2022}.
\newblock \bibinfo{title}{Centerformer: Center-based transformer for 3d object detection}, in: \bibinfo{booktitle}{Eur. Conf. Comput. Vis.}

\end{thebibliography}


\end{document}